\documentclass[letterpaper]{article} 
\usepackage{aaai2027}  
\usepackage{times}  
\usepackage{helvet}  
\usepackage{courier}  
\usepackage[hyphens]{url}  
\usepackage{amssymb}
\usepackage{graphicx} 
\usepackage{natbib}  
\usepackage{caption} 
\usepackage{amsmath}
\usepackage[table]{xcolor}

\definecolor{contourNavy}{HTML}{292742}
\definecolor{contourDeepPurple}{HTML}{513B8C}
\definecolor{contourViolet}{HTML}{8446D8}
\definecolor{contourMagenta}{HTML}{C43788}
\definecolor{contourCrimson}{HTML}{E52E5D}
\definecolor{contourOrange}{HTML}{F4772B}
\definecolor{contourGold}{HTML}{F9AA24}
\definecolor{contourYellow}{HTML}{F8D26A}
\definecolor{contourMist}{HTML}{F8FAFC}

\usepackage{algorithm}
\usepackage{algorithmic}
\usepackage{booktabs}
\usepackage{newfloat}
\usepackage{listings}
\DeclareCaptionStyle{ruled}{labelfont=normalfont,labelsep=colon,strut=off} 
\floatstyle{ruled}
\newfloat{listing}{tb}{lst}{}
\floatname{listing}{Listing}

\title{RRFC: Recursive Refinement via Feedback Conditioning for \\ Iterative Image-to-Image Generation}
\author{
    Kareem Hassani,
    Chaymaa Abbas,
    Hadi Al Mubasher,
    Mariette Awad
}
\affiliations{
    Department of Electrical and Computer Engineering,\\
    American University of Beirut,\\
    Riad El-Solh, Beirut 1107 2020, Lebanon\\
    \{krh10, cwa07, hma154\}@mail.aub.edu, mariette.awad@aub.edu.lb
}

\begin{document}

\maketitle

\begin{abstract}
Conditional image-to-image generators are single-shot: they map input features to an output in one forward pass and treat it as final, with no opportunity to improve on it. Although trained to produce the best possible result in one step, such a model leaves room for improvement if it can adaptively revise its own output over iterations. We propose Recursive Refinement via Feedback Conditioning (RRFC), a novel feedback-conditioning framework for iterative output refinement that teaches a model to adaptively revise its output by conditioning on a new signal, namely its most recent previous prediction, which is fed back as an auxiliary set of channels alongside the original input. This preserves the generator's core architecture while modifying its conditioning interface and, depending on the model family, its training or inference procedure, so RRFC can be attached to existing generators without redesign. We evaluate RRFC across six baselines spanning adversarial, equilibrium, and diffusion-based models and three paired image-to-image translation tasks. Across 18 architecture-task settings, RRFC yields seven Holm-corrected improvements, seven degradations, and four non-significant changes. The gains concentrate on reconstruction-fidelity and identity settings, while five of the seven degradations fall on the single semantic-layout task, where every model declines. These results indicate that feedback-based refinement helps when its objective overlaps with the evaluated property, and that its gains concentrate on the tasks where that overlap holds.
\end{abstract}


\section{Introduction}
\label{sec:introduction}

Most conditional generative models operate as one-shot systems. They take an input condition, produce a single output, and treat that output as final. Correction happens only offline, through backpropagation on model parameters, so improvements occur statistically across the training distribution rather than on any specific sample. Once an image is generated, there is no structured mechanism for the model to look at its own output, recognize local errors, and refine the result in a targeted way.

A natural way to add such a mechanism is to let a model condition on its own previous output and generate again. If the model can read what it just produced, it has the chance to notice where it fell short and revise that specific sample rather than relying only on what was learned across the whole training set. This idea of conditional self-refinement has appeared in several forms, such as iterative generators \citep{itergans}, self-conditioned diffusion \citep{analogbits}, and repeated refinement in super-resolution \citep{sr3}. In each of these cases the refinement is built into one specific model as part of its architecture and is designed for one task. Our proposal takes a different route.

\begin{figure}[t]
\centering
\includegraphics[width=\linewidth]{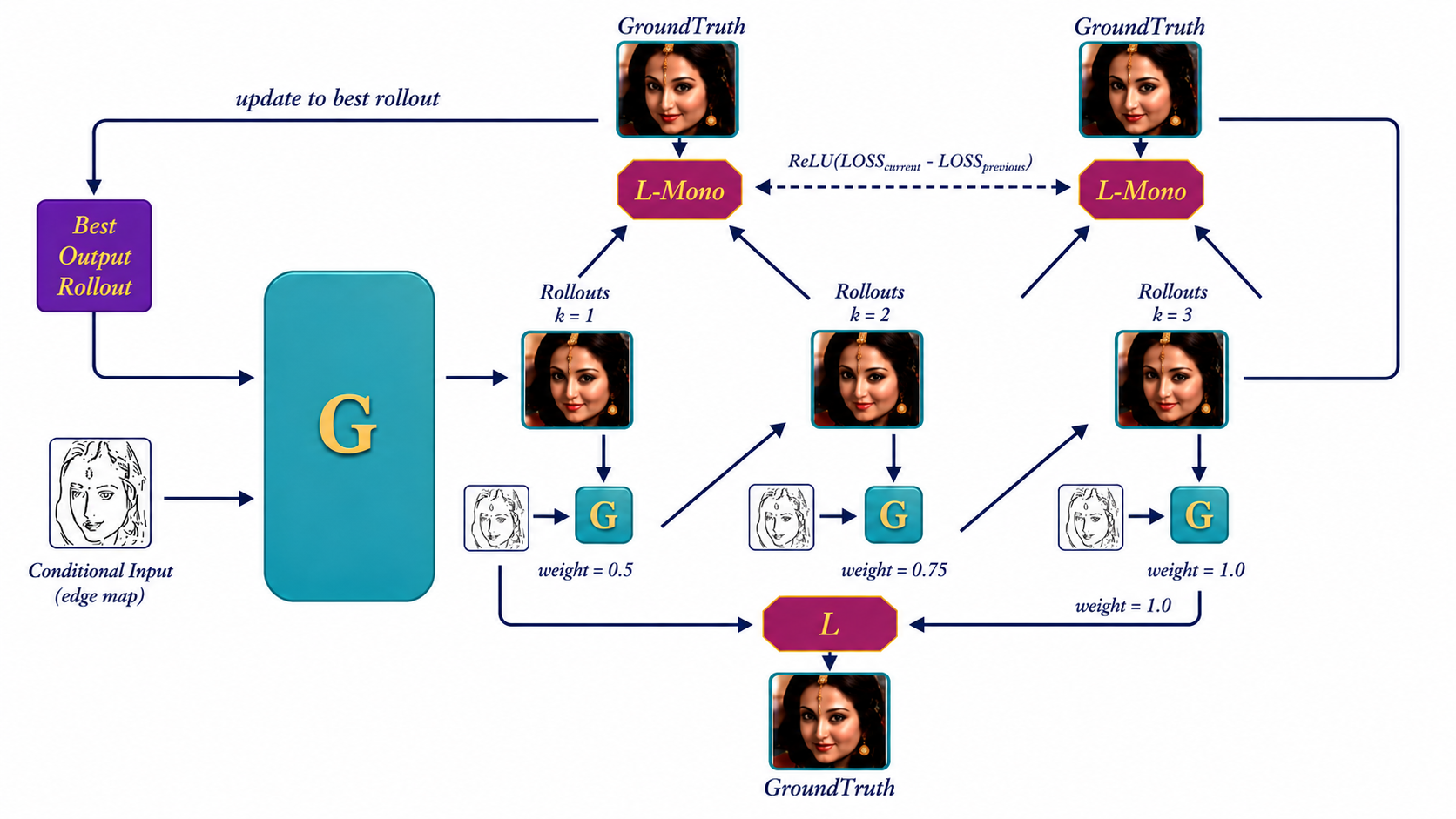}
\caption{Overview of RRFC. A generative model learns to refine each sample by conditioning on its previous prediction alongside the original input.}
\label{fig:block}
\end{figure}

We propose Recursive Refinement via Feedback Conditioning (RRFC), a feedback-conditioning framework that teaches a model to revise its own output. Rather than building refinement into the structure of one model, RRFC feeds a model's own previous prediction back as extra input channels alongside the original condition and trains the model to read that feedback and improve on it, so refinement is learned as a behavior rather than wired into the architecture. RRFC preserves the generator's core architecture and modifies its conditioning interface, together with either its training or, for internally iterative models, its inference procedure, so the same principle attaches to many generators without significant redesign. This separates RRFC from prior self-refinement, which is tied to one architecture and one task (Section~\ref{sec:method}).

To understand the effect of this intervention, we compare each model with and without RRFC on six baselines spanning adversarial, equilibrium, and diffusion-based generators and on three paired image-to-image tasks, across three seeds (6 architectures $\times$ 3 tasks $\times$ 2 variants $\times$ 3 seeds). Each output is scored on general image quality metrics that apply across all tasks and on one task-specific metric measuring the property that task actually cares about. Because these targets differ from task to task, our aim is not to improve every metric at once, which would be an overreaching claim, but to establish where self-refinement helps: most where the target is a property that revising an output can directly change, such as reconstruction fidelity and identity preservation, and not where the target is semantic layout.

The contributions of this paper are the following.

\begin{itemize}
    \item \textbf{Principle.} We present RRFC as an architecture-adaptable framework for self-refinement, rather than a single fixed mechanism, so that it can be attached to models that generate in one pass and to models that generate through an internal iterative process. This is defined in Section~\ref{sec:method}.
    \item \textbf{Study.} We run a controlled comparison of each baseline with and without RRFC across six architectures, three tasks, and three seeds, using paired significance testing with multiple comparison correction. The setup is described in the Experimental Setup (Section~\ref{sec:setup}).
    \item \textbf{Findings.} We show that whether RRFC helps a given setting is governed largely by how well the refinement objective overlaps with the metric used to evaluate that task: gains are statistically significant on reconstruction-fidelity and identity settings, while the semantic-layout task declines in every model. Responsiveness also varies with architecture, and the one baseline that already refines through an internal solve is unresponsive. These results are presented in Section~\ref{sec:results}.
\end{itemize}

\section{Related Work}

Refinement over multiple passes is an established route to higher quality in conditional generation, but it has almost always been realized as a property of one architecture built for one task. Cascaded refinement networks synthesize a photograph from a semantic layout through a fixed coarse-to-fine stack \citep{crn}, stacked adversarial models rectify the defects of an earlier low-resolution stage \citep{stackgan}, iterative adversarial generators call a single generator repeatedly to advance a controlled transformation \citep{itergans}, and diffusion-based super-resolution treats repeated stochastic denoising as the generative process itself \citep{sr3}. In each case the number of passes, the objective, and the interface between passes are fixed at design time, so refinement cannot be separated from the model carrying it. RRFC separates the two, specifying what a generator does with its previous output while leaving the generator unchanged.

Conditioning a model on its own prediction is the mechanism closest to our self-conditioning component. Diffusion samplers reuse the previous estimate of the clean image at each denoising step \citep{analogbits}, and latent self-conditioning warm-starts a set of latent tokens with state carried over from prior computation \citep{rin}. Both operate inside a single sampler. The feedback in RRFC is instead a completed output returned as image channels between full generations, making it available to models whose forward pass has no internal trajectory, and it is paired with an objective constraining how successive outputs relate.

Training on feedback the model does not produce at test time reintroduces the mismatch identified in sequence prediction, where a model supervised on ground-truth history is evaluated on its own predictions \citep{scheduledsampling}. The same mismatch has been characterized inside diffusion sampling, where the network encounters its own accumulated error at inference and corrections are applied by rescaling the prediction or shifting the sampling schedule \citep{epsscaling,expbiasshift}. Our exposure-matched feedback component acts on the feedback distribution rather than the schedule: the mixture of blank, fresh, and buffered feedback is set so that the conditioning inputs seen during training span the range encountered at inference.

Whether a model already refines internally is a property RRFC uses predictively rather than one it inherits. Equilibrium models drive a weight-tied layer to a fixed point \citep{deq}, an approach extending to diffusion sampling itself \citep{deqdiffusion}, so both families already spend inference computation on internal iteration and an added outer loop duplicates it. That cost connects to work on allocating computation at inference, from learned halting \citep{act} to search over sampling trajectories \citep{ttscaling}. RRFC exposes refinement depth as an explicit inference parameter selected on validation data, so computation spent can be read against quality returned.

Non-decreasing quality has been treated as a surrogate constraint in policy optimization \citep{trpo}, as a post-hoc construction making an early-exit classifier's accuracy conditionally monotone in depth \citep{jazbec}, and, in recent generative work, as an emergent property of a larger sampling budget rather than an enforced constraint \citep{tim}. Two recent proposals share part of our setting without occupying it: refinement as a plug-in training paradigm for autoregressive generation, where overlapping token windows revise earlier content \citep{tensorar}, and as an inference-time loop where an external critic drives repeated regeneration for compositional prompts \citep{iterrefine}. Neither penalizes a pass for being worse than the pass before it. To our knowledge, no prior image-generation work applies a pass-to-pass monotonic penalty across completed generations, which is the constraint that allows refinement depth to extend past the depth used in training.

Our baselines span the two families defined above. Pix2Pix \citep{pix2pix}, Pix2PixHD \citep{pix2pixhd}, and SAGAN \citep{sagan} generate in one forward pass, while DEQ \citep{deq}, ControlNet \citep{controlnet}, and Palette \citep{palette} generate through an internal iteration. Tasks follow the standard protocols for Cityscapes \citep{cityscapes}, Places365 \citep{places365}, and CelebA-HQ \citep{celebamaskhq}, scored with SegFormer \citep{segformer}, ArcFace \citep{arcface}, LPIPS \citep{lpips}, FID \citep{fid}, and KID \citep{kid}.

\section{Methodology: RRFC}
\label{sec:method}

RRFC turns a standard conditional generator into one that reads its own previous output and tries again. We state the principle, formalize the four components, and explain how they adapt across models.
\begin{table*}[!tbp]
\centering

\begin{minipage}[t]{0.52\textwidth}
\centering
\setlength{\tabcolsep}{7pt}
\renewcommand{\arraystretch}{1.2}

\begin{tabular}{l cc}
\toprule
\rowcolor{contourNavy}
\color{white}\textbf{Component}
&
\cellcolor{contourDeepPurple}\color{white}\textbf{Single pass}
&
\cellcolor{contourMagenta}\color{white}\textbf{Internal iter.}
\\
\midrule

\rowcolor{contourDeepPurple}
\color{white}C1 feedback to generator
& \color{white}Yes
& \color{white}Yes \\

\rowcolor{contourDeepPurple}
\color{white}C4 exposure matched feedback
& \color{white}Yes
& \color{white}Yes \\

\midrule

\rowcolor{contourOrange}
C1 feedback to discriminator
& Yes
& n.a.$^{\dagger}$ \\

\rowcolor{contourOrange}
C2 unrolled supervision
& train
& infer \\

\rowcolor{contourGold}
C2 gradient through passes
& Yes
& No \\

\rowcolor{contourGold}
C3 monotonic comparison
& pass
& fb/none \\

\rowcolor{contourYellow}
C3 weight $\lambda_m$
& 50
& 1.0$^{\ddagger}$ \\

\bottomrule
\end{tabular}

\captionof{table}{RRFC components by model family. Purple rows are identical everywhere; orange and gold rows adapt to single pass versus internal iteration. \emph{train}/\emph{infer}: refinement in training versus inference. \emph{pass}: penalty compares successive passes; \emph{fb/none}: feedback versus no feedback. $^{\dagger}$DEQ keeps an optional unconditioned discriminator; the diffusion models have none. $^{\ddagger}$The internally iterative column reports the value used by the two diffusion bases; DEQ uses $\lambda_m = 50$, because its penalty is measured in the same reconstruction space as the single-pass bases, as explained under C3 below.}
\label{tab:realization}
\end{minipage}%
\hspace{0.03\textwidth}%
\begin{minipage}[t]{0.31\textwidth}
\centering
\setlength{\tabcolsep}{5pt}
\renewcommand{\arraystretch}{1.3}

\begin{tabular}{l ccc}
\toprule
\rowcolor{contourNavy}
\color{white}\textbf{Model}
& \color{white}\textbf{T1}
& \color{white}\textbf{T2}
& \color{white}\textbf{T3} \\
\midrule

\rowcolor{contourDeepPurple}
\color{white}Pix2Pix
& \color{white}15
& \color{white}15
& \color{white}15 \\

\rowcolor{contourViolet}
\color{white}Pix2PixHD
& \color{white}14
& \color{white}14
& \color{white}14 \\

\rowcolor{contourMagenta}
\color{white}SAGAN
& \color{white}8
& \color{white}8
& \color{white}8 \\

\rowcolor{contourCrimson}
\color{white}DEQ
& \color{white}4
& \color{white}4
& \color{white}4 \\

\rowcolor{contourOrange}
ControlNet
& 2
& 2
& 2 \\

\rowcolor{contourGold}
Palette
& 16
& 16
& 16 \\

\bottomrule
\end{tabular}

\captionof{table}{Experiment grid over six models and three tasks, with batch size in each cell. Every cell is run for three seeds, giving 54 runs before and 54 after.}
\label{tab:grid}
\end{minipage}

\end{table*}
\subsection{The Refinement Principle}
\label{subsec:principle}

Let $G$ map an input condition $c$ to an output image. A standard generative model produces a single output and stops. RRFC instead defines a short sequence in which each output conditions on the previous one,
\begin{align}
x^{(0)} &= G(c, \varnothing), \\
x^{(k)} &= G\!\left(c,\, x^{(k-1)}\right), \quad k = 1, \dots, K,
\end{align}
where $\varnothing$ is a blank signal on the first pass. We call $k$ the refinement depth. The principle is that reading the previous output should make the result better and never worse. With a quality measure $q$ where lower is better, training encourages
\begin{equation}
q\!\left(x^{(k)}\right) \;\le\; q\!\left(x^{(k-1)}\right),
\end{equation}
combining a pixel and a perceptual term,
\begin{equation}
q(x) \;=\; \lVert x - y \rVert_1 \;+\; \beta \,\phi(x, y),
\end{equation}
with target $y$, perceptual distance $\phi$, and weight $\beta$. This choice of $q$ is not fixed by the method; any differentiable quality measure the base already exposes can play this role, such as an adversarial realism term or the denoising objective, so $q$ is matched to each architecture rather than imposed on it.

\subsection{The Four Components}
\label{subsec:components}
\paragraph{C1: Self conditioning.}
We widen the generator input so it also receives the previous output as extra channels, blank on the first pass. The buffered feedback is detached from the computation graph and the buffer itself is not optimized, so the model reacts to its previous output rather than reshaping it. A per sample buffer stores each sample's most recent output across epochs as the training time estimate of $x^{(k-1)}$.

\paragraph{C2: Unrolled supervision.}
We unroll the rollout for $U$ passes during training and supervise each pass against $y$, weighting later passes more,
\begin{equation}
\mathcal{L}_{\text{unroll}} \;=\; \sum_{s=1}^{U} w_s \, \mathcal{L}\!\left(x^{(s)}, y\right),
\qquad w_1 \le \dots \le w_U .
\end{equation}
Where the base allows it, each pass is generated from the previous output without detachment, so gradients propagate backward through the rollout. The previous output entering each pass's reconstruction loss, and the earlier quality term in the monotonic penalty, are detached, so a pass is trained to improve on a fixed previous output rather than to alter it. We use $U = 3$, selected by the hyperparameter search described under Configuration Selection in Section~\ref{sec:setup}.

\paragraph{C3: Monotonic objective.}
\label{C3}
This is the novel component of RRFC. It penalizes any pass worse than the one before,
\begin{equation}
\mathcal{L}_{\text{mono}} \;=\; \sum_{s=2}^{U} \operatorname{ReLU}\!\left(q\!\left(x^{(s)}\right) - q\!\left(x^{(s-1)}\right)\right),
\end{equation}
giving the full objective
\begin{equation}
\mathcal{L} \;=\; \mathcal{L}_{\text{unroll}} \;+\; \lambda_m \, \mathcal{L}_{\text{mono}} ,
\end{equation}
with $\lambda_m = 50$ on the single-pass and DEQ bases and $\lambda_m = 1.0$ on the diffusion bases, as specified in Table~\ref{tab:realization} and fixed by the search in Section~\ref{sec:setup}. The weight is per family because it regularizes a different base objective in each: on the single-pass and DEQ models the penalty joins a pixel and perceptual reconstruction loss and is measured in that space, whereas on the diffusion models it joins a noise-prediction objective on a different scale. A shared weight would impose a different effective constraint on each family; the smaller diffusion value prevents the monotonic term from dominating, rather than reflecting a weaker constraint.

\paragraph{C4: Exposure matched feedback.}
Since inference feedback is a fresh output but buffered feedback can be stale, each batch draws feedback from a mixture of three sources, namely a blank signal 15 percent of the time, a freshly generated output 55 percent, and the stored buffer the remaining 30 percent. This matches training to the range of feedback seen at test time. These proportions, in particular the 55 percent fresh exposure, are the values chosen by the hyperparameter search in Section~\ref{sec:setup}.

\paragraph{Selection.}
RRFC adds one inference choice, the number of refinement passes. To select it we roll a trained model out beyond the training depth at each epoch, sweeping passes from zero to ten and scoring every depth on the validation set. The reported checkpoint and operating depth $k$ are those giving the best validation score.

\subsection{Realization Across Architectures}
\label{subsec:realization}

How two components are realized depends on the base model, as summarized in Table~\ref{tab:realization}. The deciding property is whether it generates in one forward pass or through internal iteration. Self conditioning and exposure matched feedback are identical everywhere. The unrolled supervision and monotonic objective adapt.

For single pass models the rollout is unrolled in training, gradients flow through the passes, and the penalty compares successive passes directly. For internally iterative models, a single output is itself a long internal loop, so unrolling several full generations with gradients through the chain would require backpropagating through many stacked internal steps at once, which exceeds memory; for the fixed point base it also conflicts with a design that avoids differentiating through its solver. Refinement is therefore applied as an outer loop at inference, and the monotonic constraint becomes that feedback should not worsen the current estimate. This form is weaker and, as our analysis shows, contributes little on the diffusion bases; we keep it for consistency and report its limited effect rather than tuning it away. Discriminator conditioning is an optional auxiliary of C1, so a model with an unconditioned discriminator is still a complete instance, and \emph{n.a.} marks models with no discriminator. Every base carries all four components in the form its architecture admits.

\subsection{Scope of the Comparison}
\label{subsec:validity}

The variation in Table~\ref{tab:realization} is confined to components that must adapt, follows one fixed rule rather than per case tuning, and differs only where the alternative is infeasible, so every model is a faithful instance of the same principle. Each model is judged against itself: the before model is the unmodified generator and the after model is the same generator with RRFC, trained on the same data, number of epochs, optimizer settings, and seeds, and evaluated identically. RRFC nevertheless adds training computation through unrolling and inference computation through repeated passes, so the two arms are matched on data and schedule but not on compute. The within pair difference therefore estimates the effect of the complete RRFC intervention, including its additional computation, and does not isolate feedback conditioning from compute; a compute-matched control, in which the baseline receives equivalent additional computation without feedback conditioning, would separate the two and is left to future work. Since both share the same reduced budget, absolute scores are not meant to match published numbers; only the paired difference is interpreted.
\section{Experimental Setup}
\label{sec:setup}
\begin{figure*}[htbp]
\centering
\includegraphics[width=0.68\textwidth]{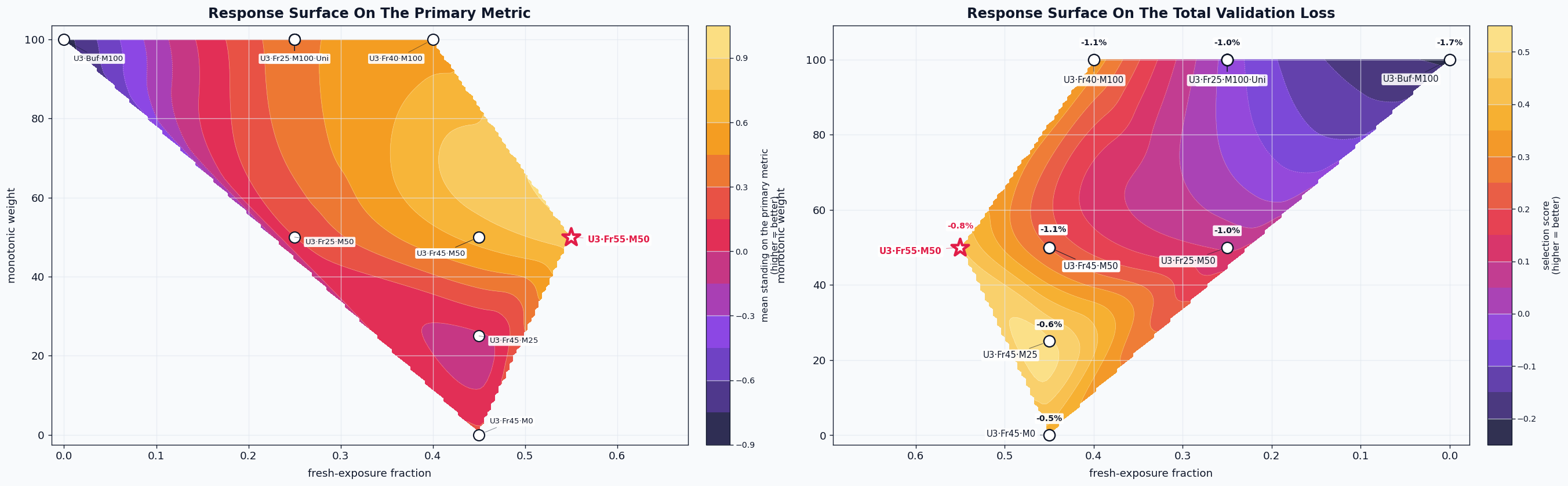}
\caption{Response surface of the RRFC hyperparameter search on the two single pass baselines. Left, standing on the reported primary metric. Right, the validation selection surrogate. The chosen global recipe (unroll 3, fresh exposure 55 percent, monotonic weight 50) is marked with a star. It is top ranked on the primary metric and within the top cluster on the surrogate.}
\label{fig:surface}
\end{figure*}

\paragraph{Models.}
We evaluate RRFC on the six baselines listed in Table~\ref{tab:realization}, spanning the two families defined in the Methodology. The single forward pass family is Pix2Pix, Pix2PixHD, and SAGAN, and the internally iterative family is DEQ, a fixed point solver, and ControlNet and Palette, two diffusion samplers. Each model is evaluated with and without RRFC under an otherwise identical pipeline, giving the eighteen model and task settings summarized in Table~\ref{tab:grid}.

\paragraph{Tasks and data.}
We use three paired image-to-image tasks, each chosen so that its target measures a different property of the output. Task T1 is mask to photo on Cityscapes, evaluated by semantic agreement of the generated photo (mIoU under a fixed SegFormer segmenter). Task T2 is inpainting on Places365, evaluated by reconstruction fidelity inside the masked region (in-hole PSNR). Task T3 is sketch to face on CelebA-HQ, evaluated by identity preservation (ArcFace cosine similarity). Each task has frozen, hash based train, validation, and test splits that are fixed once and reused for every model and seed. From these we draw a fixed working subset of 1500 training, 500 validation, and 200 test images per task. The only exception is T1 validation, which uses 347 of the 500 images in the published Cityscapes validation split after. 

\paragraph{Metrics.}
Each task has one primary metric, named above, that measures its specific target. Alongside it we report general image quality metrics that apply across all tasks, namely LPIPS, PSNR, SSIM, FID, and KID. KID is our headline distributional metric. The usual KID estimator is unbiased for squared MMD, although its variance can remain substantial with only 200 test images, so we read KID changes as directional evidence rather than as precise effect magnitudes.

\paragraph{Training protocol.}
Every model, with and without RRFC, is trained under the same budget, namely 1500 training and 500 validation images per task, except for the 347 eligible T1 validation images, for 15 epochs with Adam and three seeds. The only per model difference is batch size, fixed by memory and constant across tasks, shown in Table~\ref{tab:grid}. Because both versions (with and without RRFC) share this reduced budget, absolute scores are not meant to match full scale published numbers, and we interpret only the paired difference between a model and its RRFC counterpart.

\paragraph{Configuration selection.}
RRFC has a small number of hyperparameters, namely the unroll depth, the fresh exposure fraction, the monotonic weight, and the pass weighting schedule. We select a single global recipe once, on validation data only, and apply it to the single-pass formulation, with a predefined architecture-family adaptation for internally iterative models, as specified in Table~\ref{tab:realization}. The search is run on two RRFC-augmented baselines (Pix2Pix and Pix2PixHD) over thirteen recipes across three tasks; Figure~\ref{fig:surface} shows the response surface. We adopt the most generalizable recipe, namely unroll depth 3, fresh exposure 55 percent, and monotonic weight 50. It ranks first on the reported primary metric and lies within the top cluster on the selection surrogate, within roughly one percent of the best on average across cells. The final recipe was selected according to the predefined validation ranking; the zero-weight alternative is retained as an ablation to quantify the monotonic component's contribution. Full search diagnostics are in the appendix, including the effect of each hyperparameter (Figure~\ref{fig:a2}), the per cell rank heatmap (Figure~\ref{fig:a3}), the closeness of the top recipes (Figure~\ref{fig:b1b}), the primary metric standing (Figure~\ref{fig:b9}), and a leave one out component ablation (Figure~\ref{fig:b7}).
\paragraph{Evaluation and statistics.}
For each setting we select the training checkpoint and the number of refinement passes on validation data, never on the test set, and evaluate at that depth. Per image metrics are averaged over seeds and compared before and after with a paired two sided Wilcoxon signed rank test, with Holm correction across the eighteen model and task settings. We report bootstrap confidence intervals on the median change. FID and KID are computed at the set level and reported without per image tests. Because the T3 primary metric requires a successful face detection and a valid embedding in both arms, images for which either arm fails this requirement are excluded pairwise, and the same retained subset is used for the before and the after arm of a given cell. 

\section{Results}
\label{sec:results}

\begin{figure}[htbp]
\centering
\includegraphics[width=0.99\columnwidth]{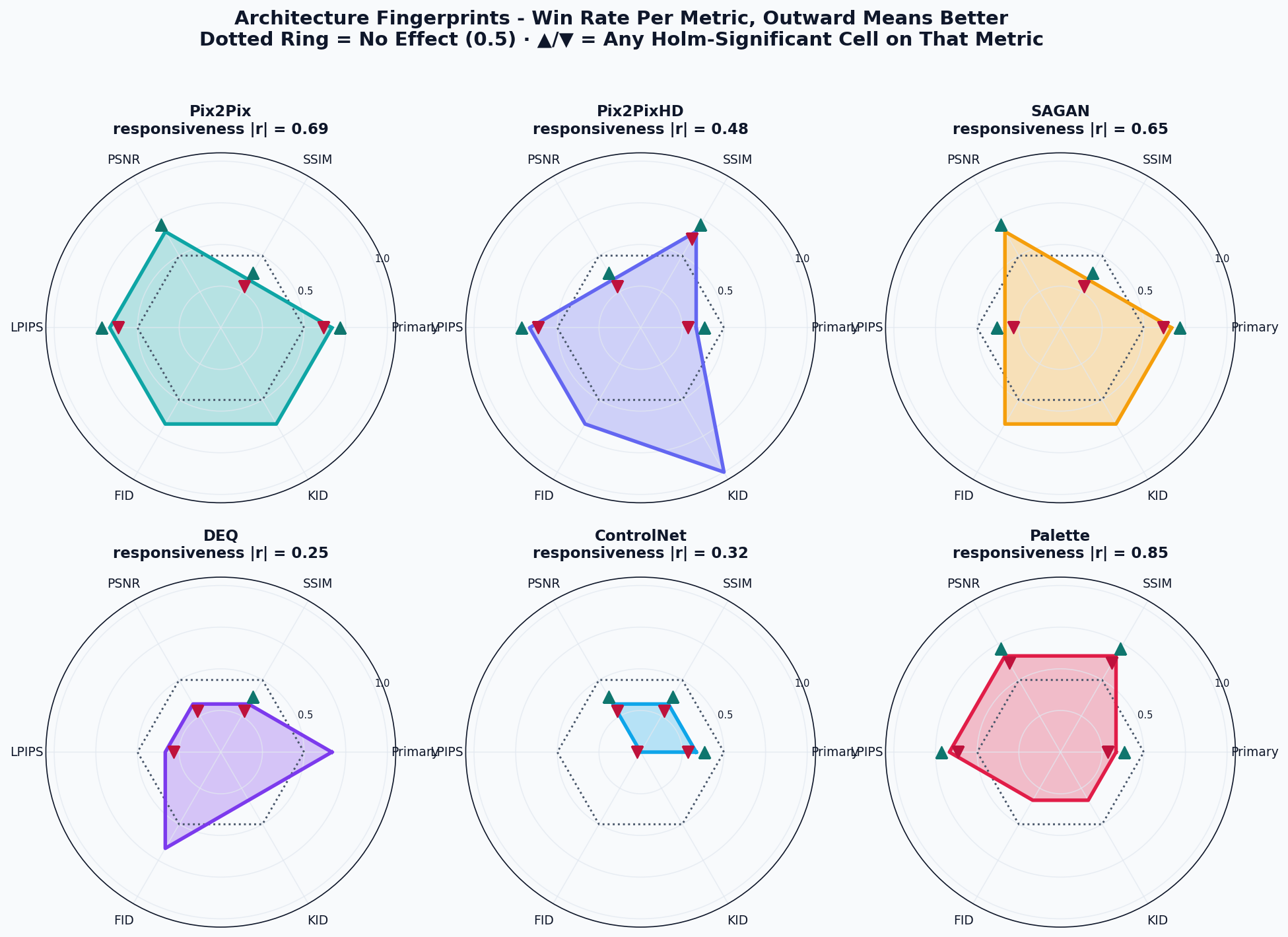}
\caption{Per model win rate across metrics, with responsiveness. DEQ, the equilibrium base, is least responsive and improves fewest metrics.}
\label{fig:radars}
\end{figure}

\begin{figure}[t]
\centering
\includegraphics[width=0.90\columnwidth]{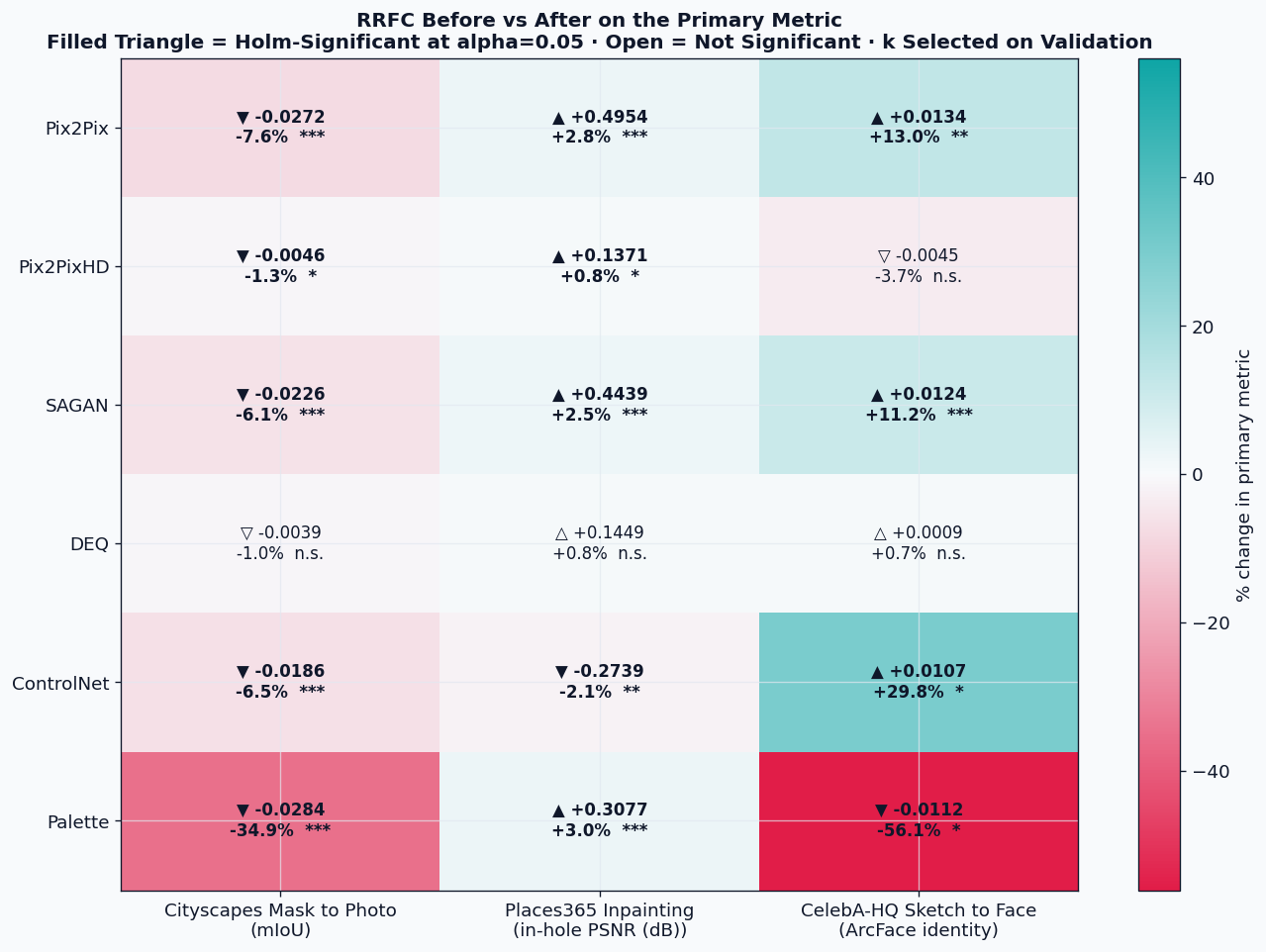}
\caption{Before and after change on each task's primary metric at the validation selected depth. Filled triangles are Holm significant at $\alpha=0.05$, open triangles are not. RRFC significantly improves seven cells, significantly degrades seven, and produces four non-significant changes, split largely by task.}
\label{fig:sigmatrix}
\end{figure}

\paragraph{Overall verdict.}
Figure~\ref{fig:sigmatrix} shows the paired before and after change on each task's primary metric, with Holm significance across all eighteen cells. RRFC produces a statistically significant improvement in seven cells, a significant degradation in seven, and no reliable change in four. The split is organized largely by task rather than architecture: five of the seven significant degradations occur on the semantic task; all six semantic-task effects are negative, with five reaching Holm-corrected significance. The remaining two significant degradations are ControlNet on inpainting and Palette on identity, while the fidelity and identity tasks account for all seven significant gains. Holm significance is a strict bar, however, and improvements that do not clear it are not thereby absent. Figure~\ref{fig:pairedbars} in the appendix shows the seed averaged primary for every cell, and on the two aligned tasks the after bars exceed the before bars in ten of twelve cells, the two exceptions being ControlNet on inpainting and Pix2PixHD on identity; the semantic task, by contrast, is negative in all six. Note also that Figure~\ref{fig:sigmatrix} reports only each task's strict primary metric. On the aligned tasks the distributional and structural metrics, KID, FID and SSIM, move positively even where the primary change is small or non-significant (Figures~\ref{fig:radars} and \ref{fig:metriccards}), which does not extend to the semantic task. The benefit on aligned tasks is therefore a broad positive tendency, of which the seven Holm significant cells are the strongest instances; the semantic loss is uniform and is analyzed in Section~\ref{sec:analysis}.

\begin{figure}[t]
\centering
\includegraphics[width=0.99\columnwidth]{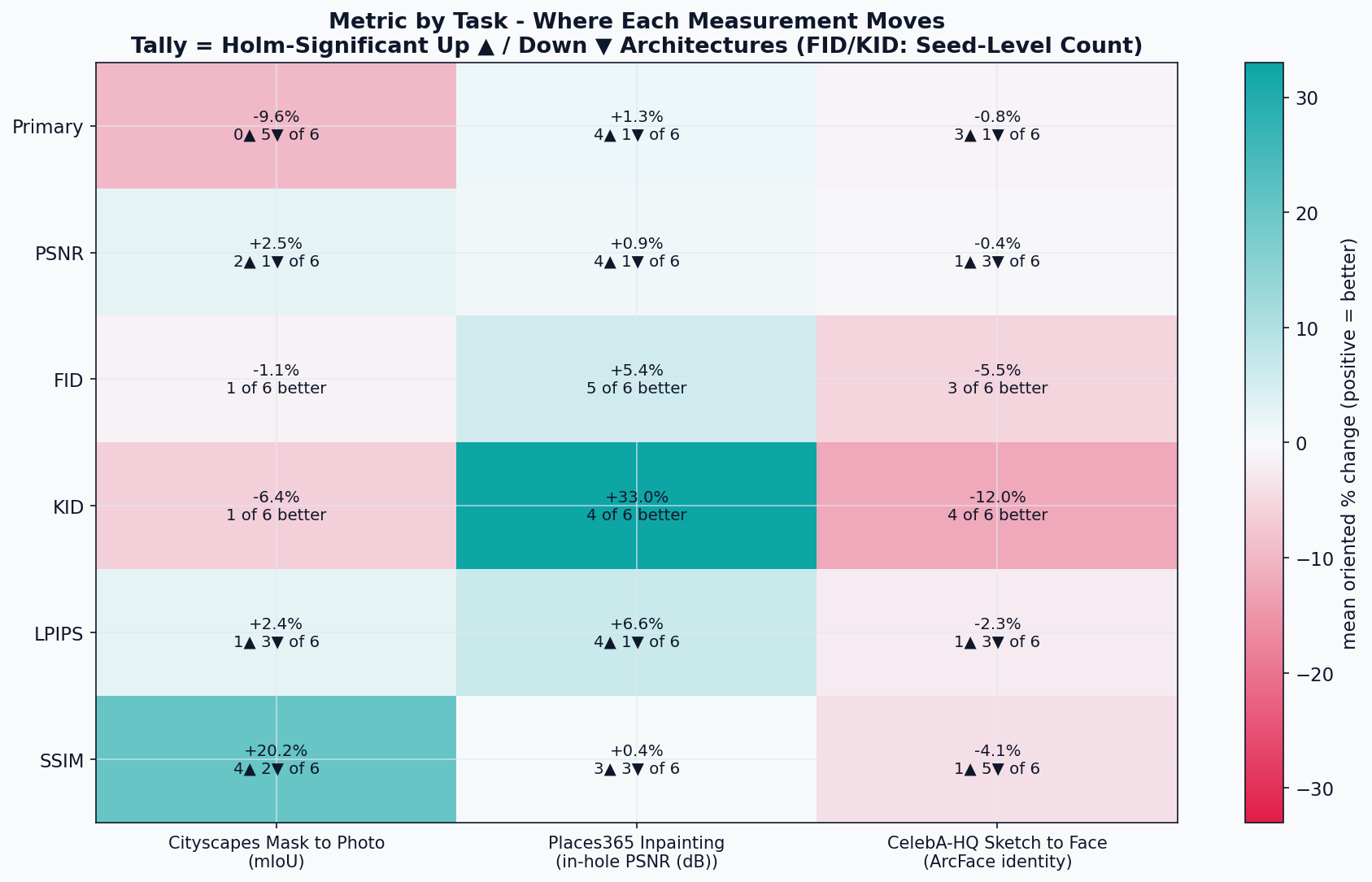}
\caption{Mean oriented change of every metric on each task, aggregated over the six models, with the count of significant models. Inpainting improves on all metrics, most strongly on KID. The semantic task improves in structural fidelity (SSIM, PSNR) while its primary metric mIoU falls.}
\label{fig:metrictask}
\end{figure}

\paragraph{Refinement improves reconstruction realism.}
The clearest benefit is on inpainting, where every metric moves in the right direction (Figure~\ref{fig:metrictask}, aggregating each metric over the six models per task). On Places365, RRFC improves in-hole PSNR (four of six models significant), and it improves every companion metric as well, namely KID by 33.0 percent, FID in five of six models, LPIPS by 6.6 percent, and SSIM. The KID change is the largest distributional movement observed in the study, and it indicates closer agreement between the generated and real image distributions in the selected feature space rather than a reduction in pixel error alone. The per model effects are large and concentrated, not outlier driven: on Pix2Pix the median per image gain is $+0.50$ dB with a matched pairs effect size $r=0.82$ and 82 percent of images improved, on SAGAN $+0.44$ dB with $r=0.76$, and on Palette $+0.31$ dB with $r=0.89$ and 86 percent of images improved. These per image distributions are shown in Figure~\ref{fig:perimage} in the appendix, and the effect sizes with bootstrap intervals in Figure~\ref{fig:forest}.

\paragraph{Refinement improves identity where identity is recoverable.}
On sketch to face, RRFC significantly raises ArcFace identity in three of six models, by 13.0 percent on Pix2Pix, 11.2 percent on SAGAN, and 29.8 percent on ControlNet (Figure~\ref{fig:sigmatrix}). Here the target is a property of one region, the face, and refinement sharpens exactly that. The absolute identity scores nevertheless remain low, near 0.10 to 0.13, below a strict same person threshold, a consequence of the reduced budget shared by both arms; the improvement is a relative gain on a difficult operating point, not a claim of solved identity.

\paragraph{Refinement is doing the work, and it is stable.}
Figure~\ref{fig:refine}, Appendix B, shows that on favorable tasks quality rises monotonically as refinement passes increase from zero and then holds flat out to ten passes, even though training used an unroll of only three. This stability beyond the training depth is consistent with the intended effect of the monotonic objective, although the leave one out ablation in Figure~\ref{fig:b7} tests that component on one architecture only. Most of the gain arrives by three to five passes, so a deployed system captures the benefit at shallow depth. The comparison is internal to the RRFC-trained models and does not by itself separate feedback conditioning from differences in training computation or first-pass behavior.

\paragraph{Where RRFC does not help.}
On Cityscapes mask to photo the primary metric falls in all six models, by 7.6 percent on Pix2Pix and up to 34.9 percent on Palette, with five of the six reaching Holm-corrected significance (Figure~\ref{fig:sigmatrix}). It is a specific failure rather than a general one, and Figure~\ref{fig:metrictask} shows why: on the same outputs, aggregate structural similarity improves by 20.2 percent and PSNR by 2.5 percent. Refinement did what its objective rewards, bringing outputs closer to the target in fidelity, but this did not raise mIoU, which measures semantic layout as read by a segmenter and is not something the fidelity objective optimizes. The semantic task is thus where metric and objective are misaligned, analyzed next.

\paragraph{Cost.}
Figure~\ref{fig:cost}, Appendix B places quality against inference latency. On the single pass GAN models, refinement is inexpensive, since each pass is one forward pass, and the useful depth of three to five passes remains far below one second per image, at the cost of three to five generator evaluations rather than one. On the diffusion models each pass is a full sampler run, so the outer loop is an order of magnitude more expensive, and where it does not help the selected depth collapses to zero, so the method declines to spend compute where it earns nothing. Latency is wall-clock GPU time for the $k$-pass refinement rollout, measured with CUDA synchronization on a single NVIDIA A100-SXM4-40GB (TensorFlow 2.20, PyTorch 2.11, CUDA 12.8) at $256\times256$ output resolution for all three tasks. Each measurement times one rollout on a batch, as is reported in Table \ref{tab:grid}. Diffusion passes use 30 DDIM sampling steps, whereas an adversarial pass is a single forward evaluation, which accounts for the order of magnitude gap between the two families. Timing covers the forward rollout only, so data preprocessing, metric computation, and device to host transfer are excluded, as is host to device transfer for the diffusion models. These figures describe inference cost only. The training cost of the unrolled objective is discussed in Section~\ref{subsec:validity}.

\section{Analysis}
\label{sec:analysis}

The results section established where RRFC helps. This section explains the pattern, identifies the property that governs it, and states what refinement does and does not change.

\paragraph{Objective-metric alignment explains the dominant pattern.}
Across tasks, effects are more favorable when the refinement objective overlaps with the evaluated property; this association is not universal at the individual architecture-task level. The objective rewards fidelity to the target, in pixels and in perceptual features. On inpainting the target is fidelity itself, so every metric improves and the primary improves in most models. On sketch to face the target is identity in the face region, which fidelity refinement can sharpen, so identity improves in the responsive models. On mask to photo the primary is semantic segmentability, which is not a fidelity quantity, and it falls, even while fidelity metrics on the same outputs rise (Figure~\ref{fig:metrictask}). The failure is therefore not that refinement broke on the semantic task, but that it optimized the property it was designed to optimize, which that task does not reward. Two observations bound this account. First, changing only the selection criterion, while holding the model, data, and test set fixed, changes the sign of the semantic effect which is the behavior a misaligned selection objective would produce. Second, alignment does not account for the two off-task degradations, ControlNet on inpainting and Palette on identity, both of which occur on aligned tasks. Alignment is therefore a task-level regularity with architecture-specific exceptions rather than a cell-level predictor, and we identify the aligned-objective test as future work in Section~\ref{sec:conclusion}.

\paragraph{Internally iterative models show mixed responsiveness.}
Figure~\ref{fig:radars} profiles each model by how many metrics it improves and how strongly it responds to RRFC at all. The responsiveness ordering is Palette at $0.85$, Pix2Pix at $0.69$, SAGAN at $0.65$, Pix2PixHD at $0.48$, ControlNet at $0.32$, and DEQ at $0.25$. DEQ is the least responsive baseline, the one base that already produces its output through an internal iterative solve, and it shows no significant change on any of the three primary metrics (Figure~\ref{fig:sigmatrix}). This is what the principle behind RRFC predicts, since a model that already iterates internally has less to gain from an added outer refinement loop. The two diffusion models, which also iterate internally through their sampler, do not follow that prediction: they exhibit mixed, task-dependent effects, with Palette the single most responsive baseline in the study and ControlNet among the least, so internal iteration alone does not determine whether an outer refinement loop will help. The single pass GAN models, which have no internal refinement, are consistently responsive and account for most of the significant improvements, so the absence of internal refinement is a reliable indicator of responsiveness while its presence is not a reliable indicator of immunity.

\paragraph{What refinement changes.}
Figure~\ref{fig:board} ranks the metrics by how much RRFC moves them across all cells. The largest movers are distributional and structural, namely KID and SSIM, with SSIM changing significantly in all eighteen cells, while full image pixel error moves least. Refinement is therefore best understood as an operation that reshapes structure and moves outputs toward the real image distribution, rather than one that reduces average pixel error. This mechanism accounts for the pattern: the inpainting gains, where structure and realism are the target; the identity gains, where sharpening the face is a structural change; and the semantic shortfall, where correct semantic layout is not something a fidelity and realism operation supplies.

\section{Discussion and Conclusion}
\label{sec:conclusion}

Self refinement, cast as an architecture-adaptable conditioning principle rather than a fixed mechanism, transfers across generators, but its benefit is predictable in aggregate rather than universal: the overlap between the refinement objective and the evaluation metric governs the dominant direction of the effect. It is thus well suited to single-pass generators on fidelity- and identity-driven tasks, where it improves the measured property at the cost of a small multiple of forward passes, but should not be expected to help on semantic metrics, on the equilibrium model, or on diffusion models, where each pass is a full sampling run. The contribution is therefore less a single win than a map of where and why iterative self refinement helps, and of the scope conditions under which that map holds.

\paragraph{Limitations and future work.}
All models are trained under one reduced budget, so we interpret paired before-after differences rather than absolute performance, and the identity gains occur from a low operating point. The arms are matched on data, epochs, optimizer settings, and seeds but not on computation, so the reported effect is that of the complete RRFC intervention and does not separate feedback conditioning from the additional computation it introduces. We select one global recipe on two GAN baselines without per-model tuning, and the monotonic term contributes little on diffusion models. Distributional comparisons emphasize KID, whose estimator variance remains substantial at 200 test images. Finally, the T3 primary metric depends on successful face detection, so affected cells are evaluated on a reduced paired subset. Future work should align selection objectives with the target metric, learn when to stop refining, evaluate under full budgets, run a compute-matched control, and strengthen the training-time monotonic constraint for diffusion models.


\bibliography{aaai2027}

\appendix
\clearpage
\appendix
\onecolumn
\section{Appendix}
\label{app:search}
\begin{figure*}[htbp]
\centering
\includegraphics[width=\textwidth]{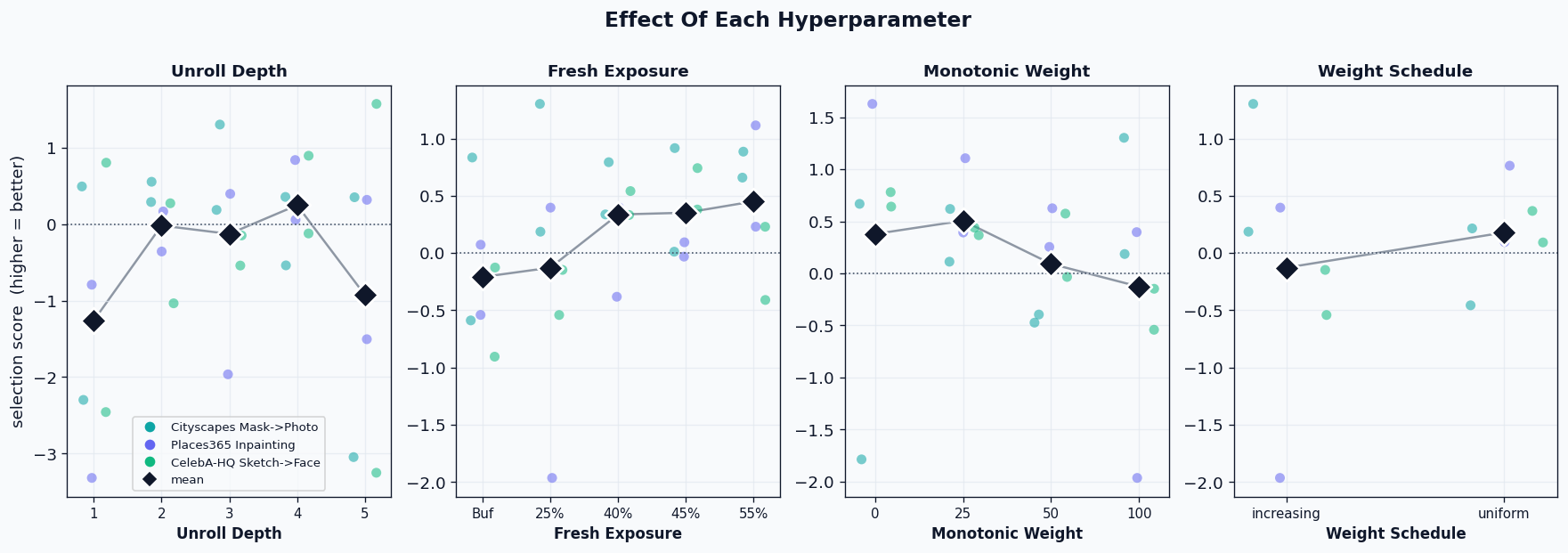}
\caption{One-factor-at-a-time view of the search: each panel varies a single hyperparameter and plots the validation selection score for every task--recipe combination (colored by task), with a black diamond at the per-setting mean. Fresh exposure shows the clearest monotone trend, rising to 55 percent, which motivates its value in the adopted recipe.}
\label{fig:a2}
\end{figure*}

\begin{figure*}[htbp]
\centering
\includegraphics[width=1\textwidth]{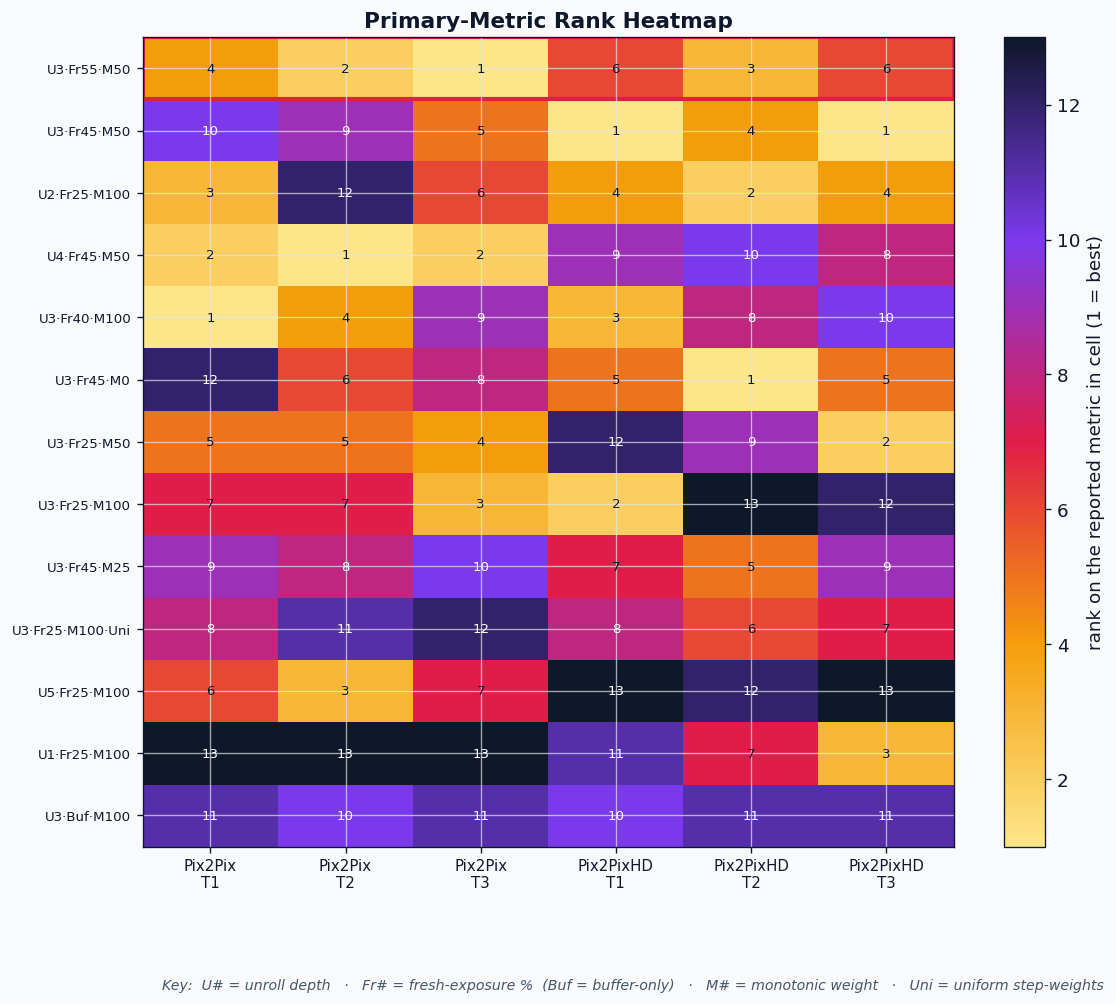}
\caption{Per-cell ranking behind the search. Rows are the thirteen candidate recipes, columns the six search cells (Pix2Pix and Pix2PixHD on T1, T2, T3), and each cell gives the recipe's rank on that task's primary metric (1 best, light; 13 worst, dark). Rows are ordered by mean rank, so the adopted recipe U3$\cdot$Fr55$\cdot$M50 sits on top: never worse than sixth and best or near-best in several, though not uniformly first. Key: U\# unroll depth, Fr\# fresh-exposure percent (Buf = buffer-only), M\# monotonic weight, Uni = uniform step-weights.}
\label{fig:a3}
\end{figure*}

\begin{figure*}[tbp]
\centering
\includegraphics[width=\textwidth]{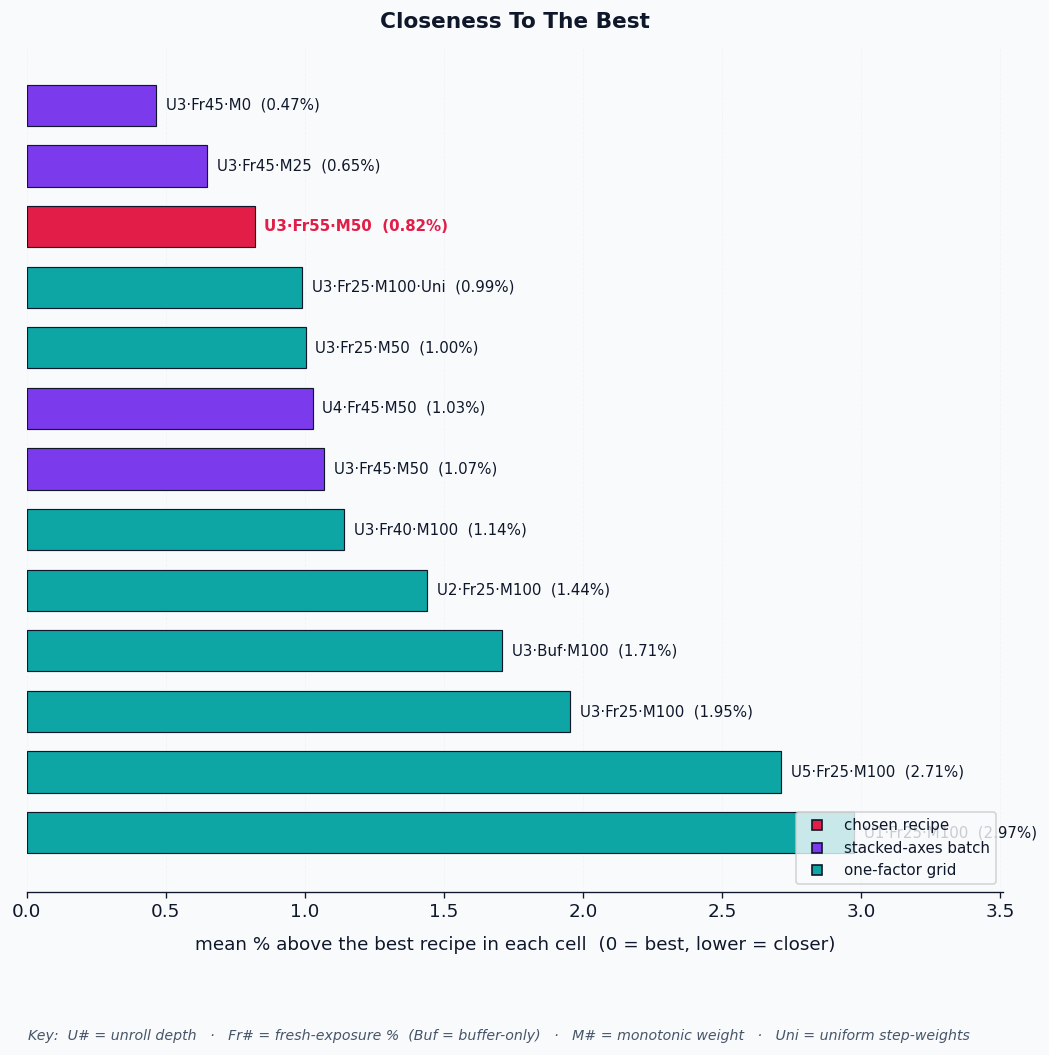}
\caption{How far each recipe sits from the per-cell best, as its mean percentage above the best recipe in each cell (0 = matches the best, lower = closer), colored by how the recipe was generated and sorted closest to farthest. The adopted recipe U3$\cdot$Fr55$\cdot$M50 is third closest at 0.82 percent, within about one percent of the cell-wise optimum. The two closer recipes carry little or no monotonic penalty; the zero-weight one is kept as the ablation for the monotonic term. Key as in Figure~\ref{fig:a3}.}
\label{fig:b1b}
\end{figure*}

\begin{figure*}[tbp]
\centering
\includegraphics[width=\textwidth]{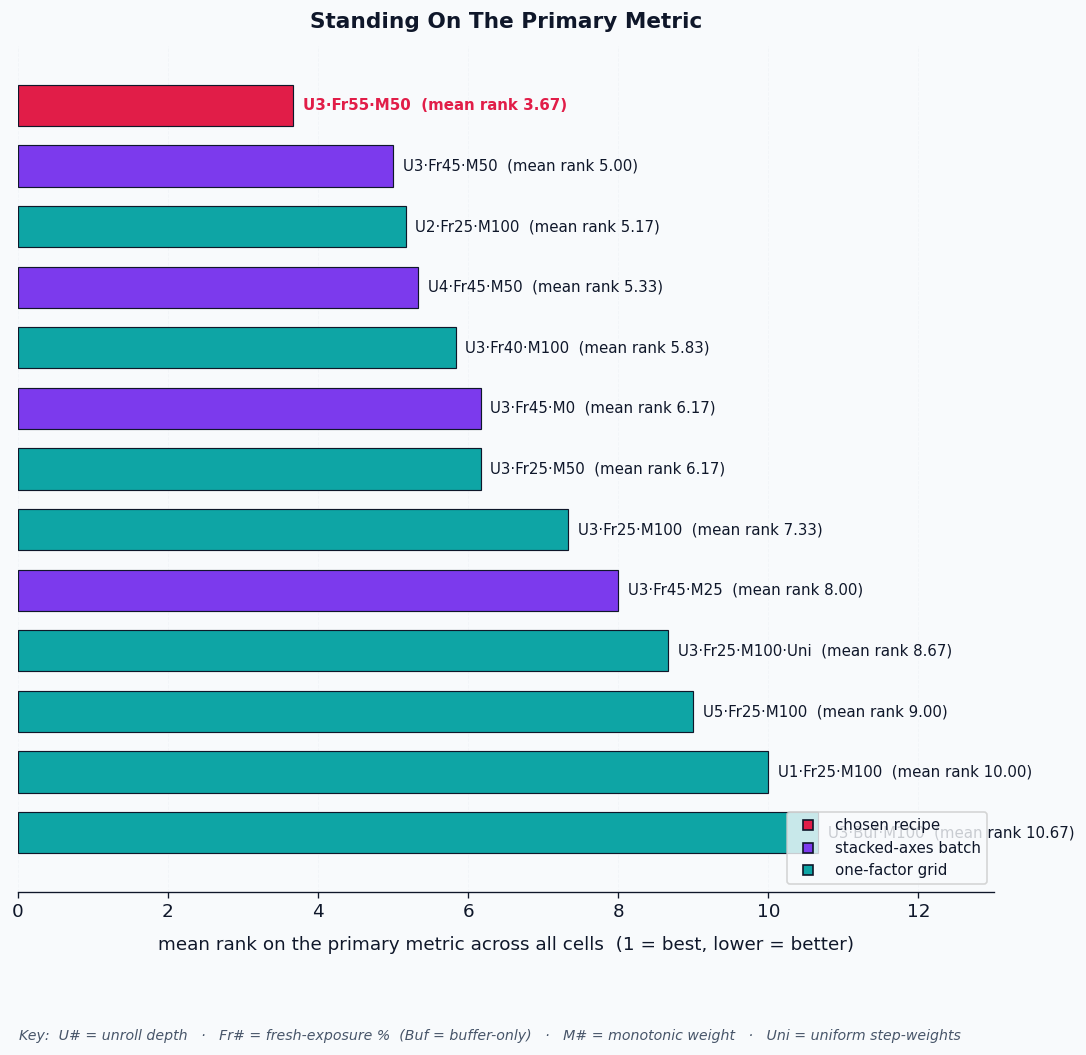}
\caption{Mean rank of each recipe on the primary metric across all search cells (1 = best, lower is better); this is the selection criterion. The adopted recipe U3$\cdot$Fr55$\cdot$M50 ranks first at 3.67, ahead of the runner-up at 5.00, so its selection does not rest on any single cell. Bar colors and key follow Figure~\ref{fig:a3}.}
\label{fig:b9}
\end{figure*}

\begin{figure*}[tbp]
\centering
\includegraphics[width=\textwidth]{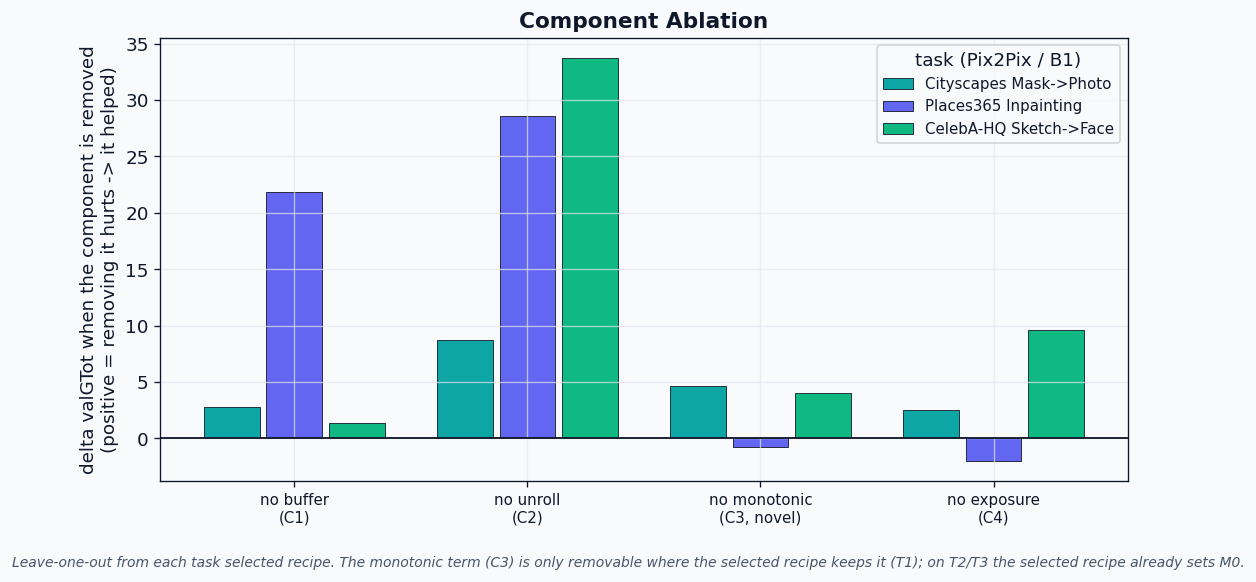}
\caption{Leave-one-out ablation of the four RRFC components on Pix2Pix, per task on that task's selected recipe. Each bar is the change in total validation loss when one component is removed, oriented so positive means removal hurts. Unrolled supervision (C2) is by far the most important, especially on the two aligned tasks; self-conditioning (C1) matters most on inpainting; exposure matching (C4) and the monotonic term (C3) have smaller, mixed effects. The monotonic term is genuinely removable only on T1, since the T2 and T3 selected recipes already set its weight to zero. Single architecture, indicative only.}
\label{fig:b7}
\end{figure*}

\begin{figure*}[tbp]
\centering
\includegraphics[width=\textwidth]{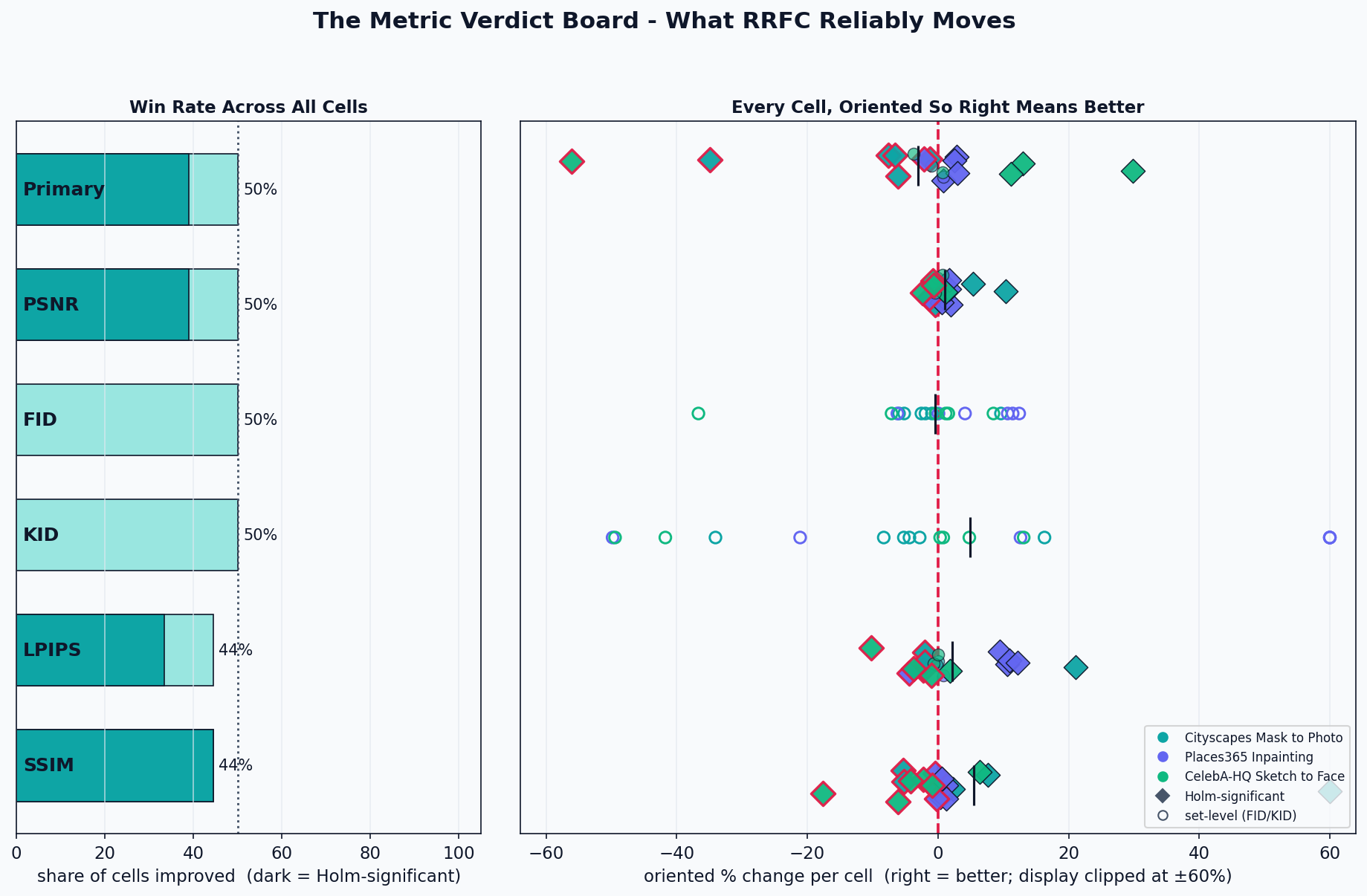}
\caption{How reliably and how far RRFC moves each metric, pooled over all eighteen cells. Left: the share of cells where the metric improves (Holm-significant share darker, 50 percent dotted); because the semantic task is pooled in, the win rates sit near a coin flip. Right: the oriented percentage change in every cell (right = better, clipped at $\pm$60 percent), filled diamonds Holm-significant and open circles the set-level FID and KID. The spread is the point: KID, then FID and SSIM, move most in both directions while pixel-level PSNR stays near zero, so refinement acts most on distribution and structure.}
\label{fig:board}
\end{figure*}

\begin{figure*}[tbp]
\centering
\includegraphics[width=\textwidth]{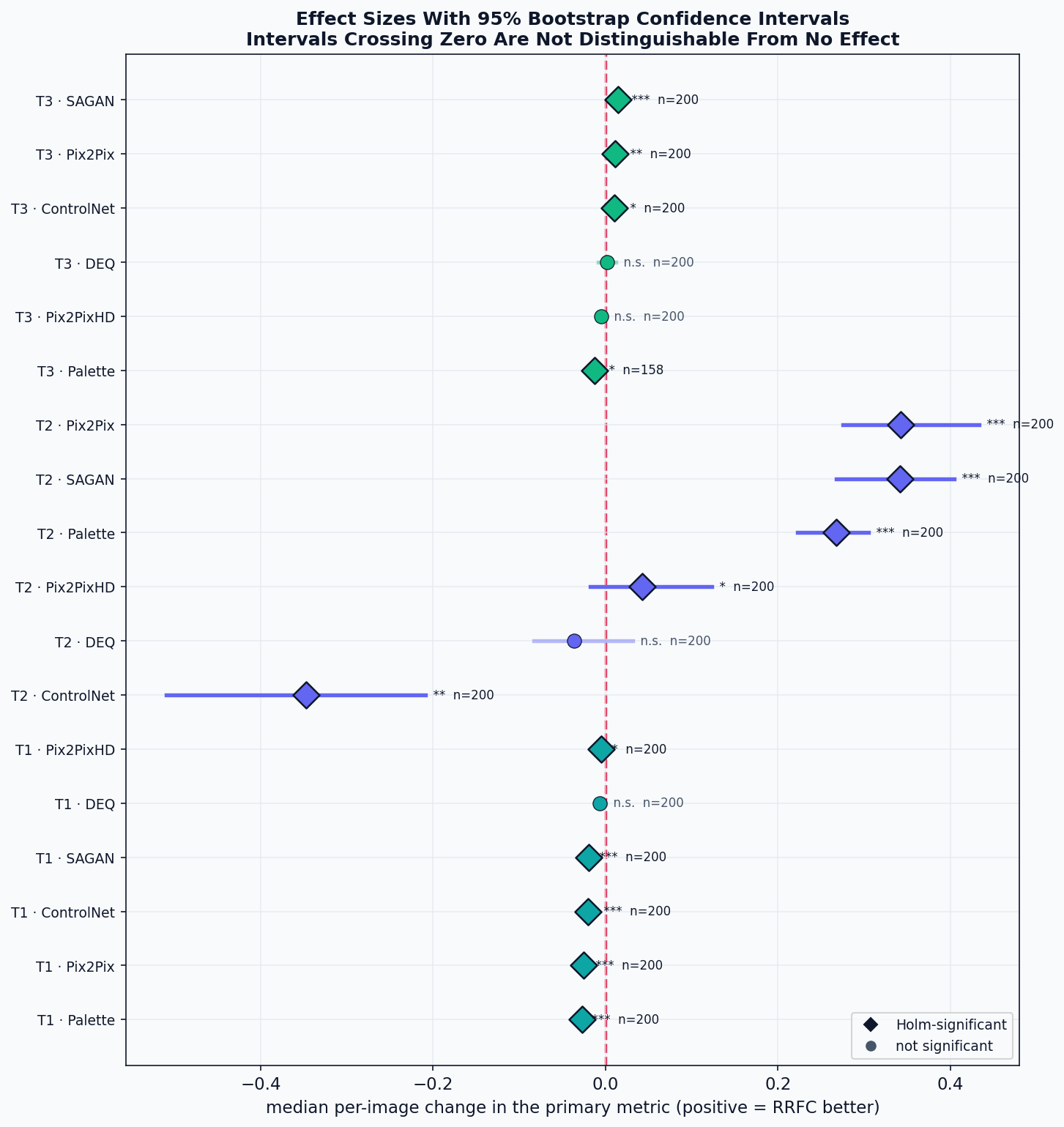}
\caption{Median per-image change in each task's primary metric across all eighteen cells, with 95 percent bootstrap intervals; any interval crossing zero is not distinguishable from no effect. Filled diamonds are Holm-significant, and $n$ labels each row (T3 Palette uses $n=158$ after face-detection failures). Inpainting (T2) carries the largest positive effects, its one strong negative being ControlNet; the semantic task (T1) is a tight band of small, mostly significant negatives; identity (T3) is small and mixed. Magnitudes are not comparable across tasks, as the primary metrics live on different scales, and because this uses the median a cell's sign can differ from the mean-based views for skewed cells.}
\label{fig:forest}
\end{figure*}

\begin{figure*}[tbp]
\centering
\includegraphics[width=\textwidth]{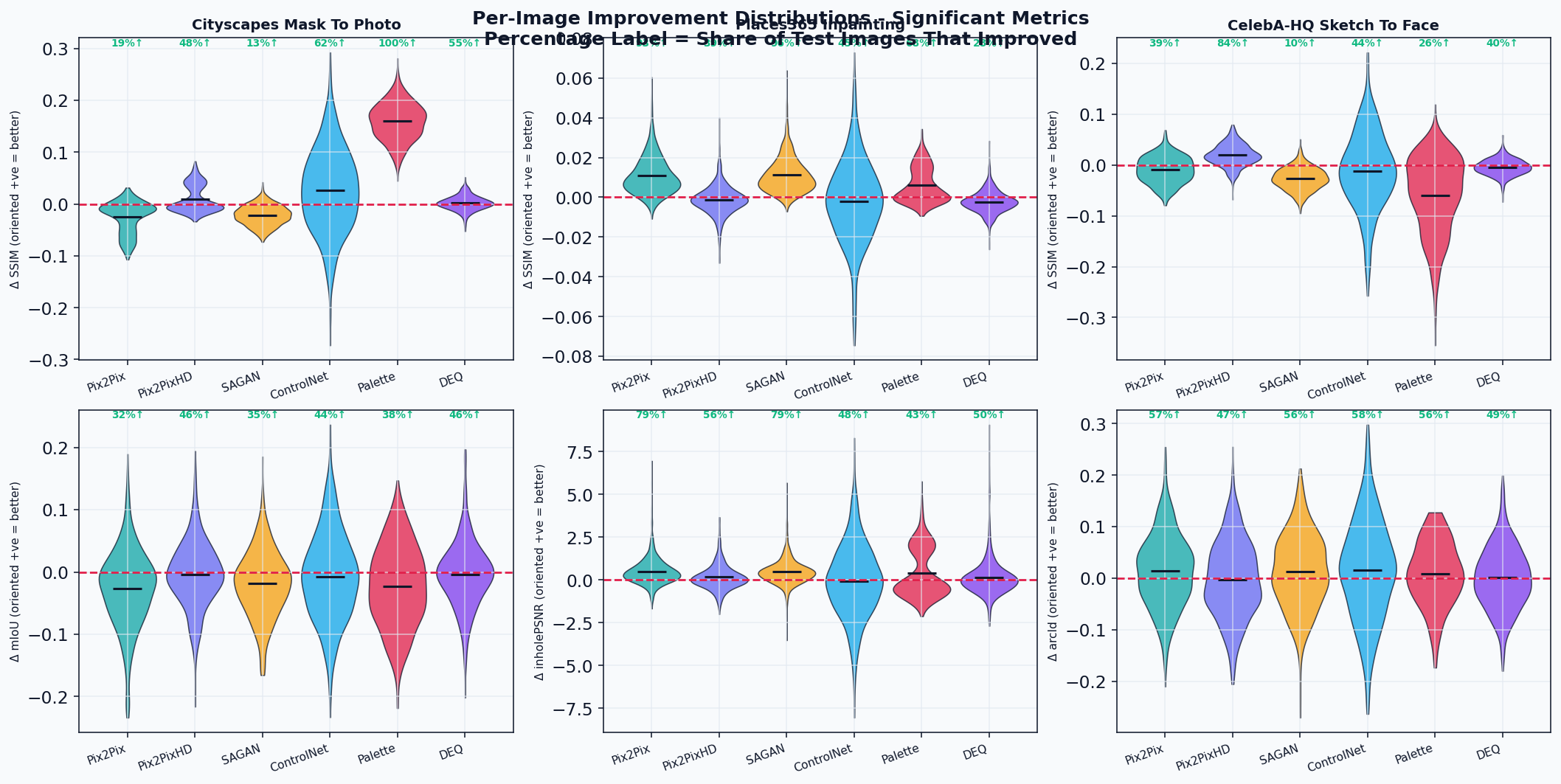}
\caption{Per-image change distributions on the significantly-moving metrics, one violin per architecture (black bar at the median), the percentage above each giving the share of images that improved. Columns are the three tasks; the top row is SSIM and the bottom row each task's primary metric. Inpainting shows the strongest per-image fidelity gains, with Pix2Pix and SAGAN improving in-hole PSNR on about four in five images. The semantic column shows the central dissociation: SSIM improves on a large share of images while the mIoU primary improves on well under half. Identity gains are broad but modest.}
\label{fig:perimage}
\end{figure*}

\begin{figure*}[tbp]
\centering
\includegraphics[width=\textwidth]{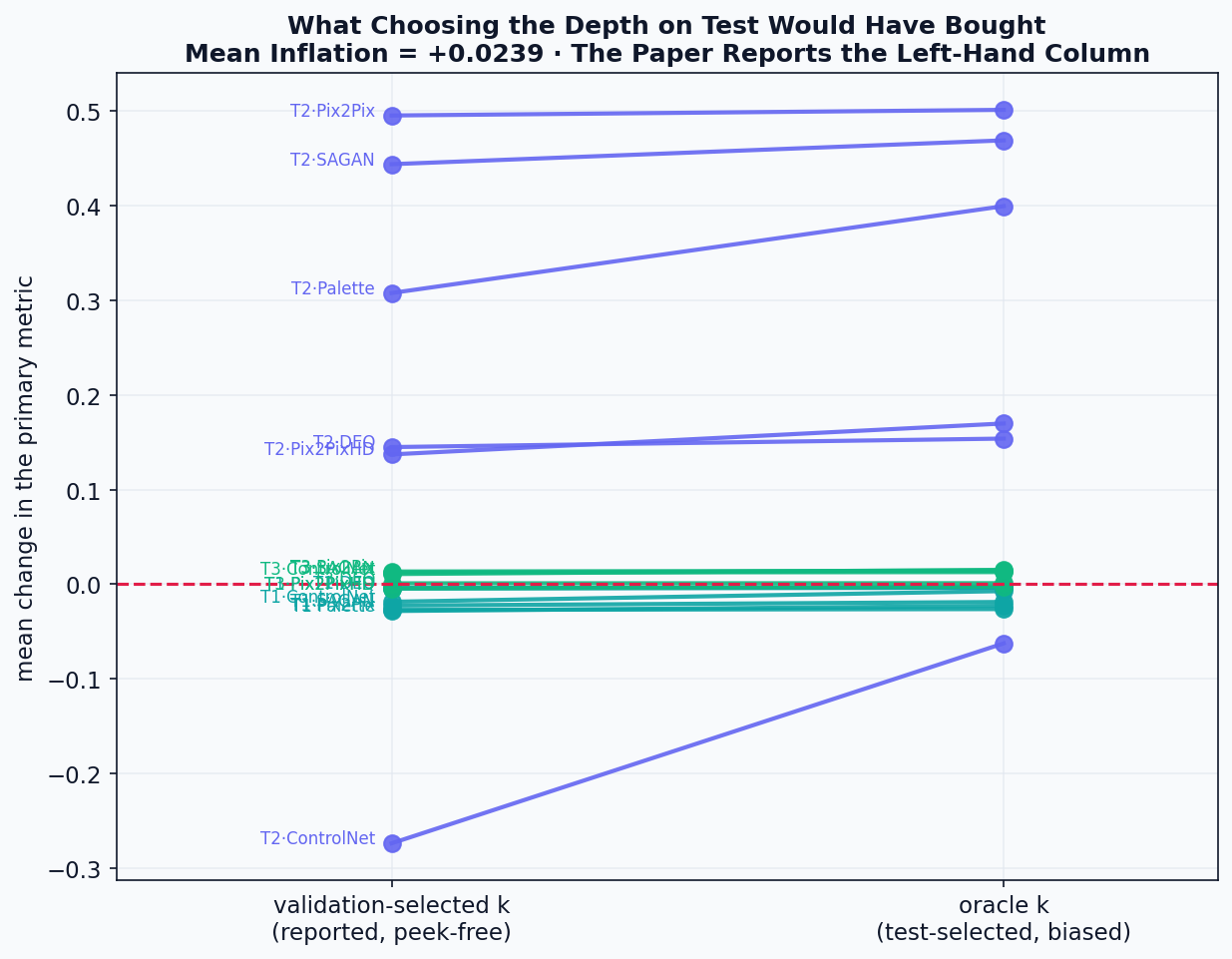}
\caption{What test-set peeking would have added. For each cell the left point is the mean primary-metric change at the validation-selected depth (reported) and the right point at the oracle depth chosen on test, joined by a line. Most lines are nearly flat and the average inflation is only $+0.024$; the largest gap, T2 ControlNet, stays negative, so that degradation is real. The reported gains do not depend on test-set access.}
\label{fig:valk}
\end{figure*}

\begin{figure*}[tbp]
\centering
\includegraphics[width=\textwidth]{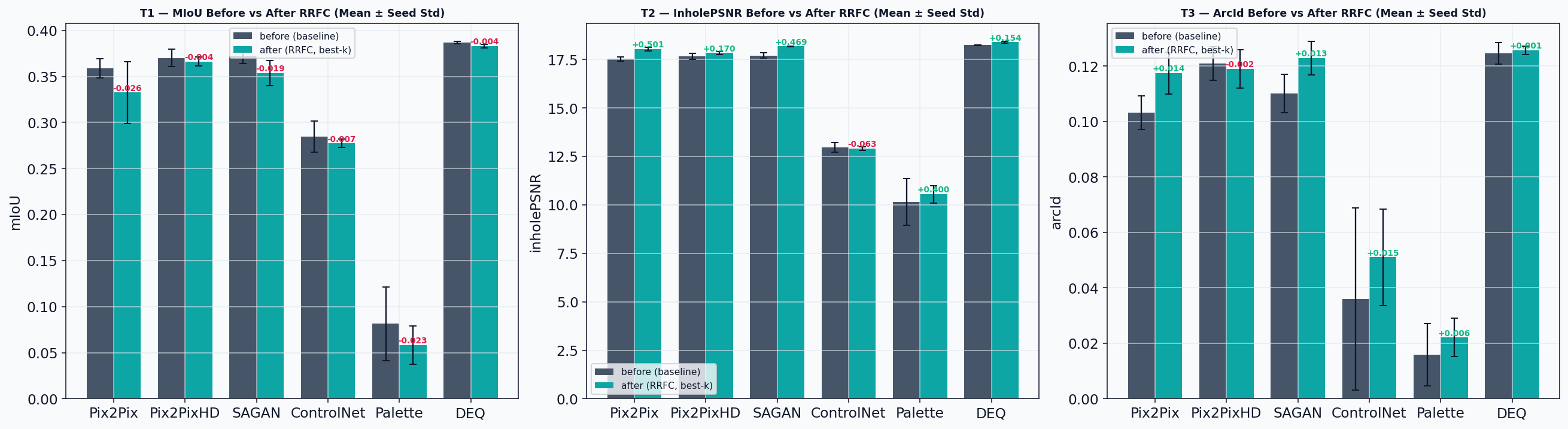}
\caption{Seed-averaged primary metric before and after RRFC, one panel per task, with seed-standard-deviation error bars and the signed change annotated. The direction is consistent within task: every model declines on the semantic task (T1); five of six improve on inpainting (T2), ControlNet excepted; most improve on identity (T3), Pix2PixHD excepted. DEQ has the highest scores and the smallest movement in this view. Many changes fall within overlapping error bars, so the value is the per-task direction rather than any single bar.}
\label{fig:pairedbars}
\end{figure*}

\begin{figure*}[tbp]
\centering
\includegraphics[width=\textwidth]{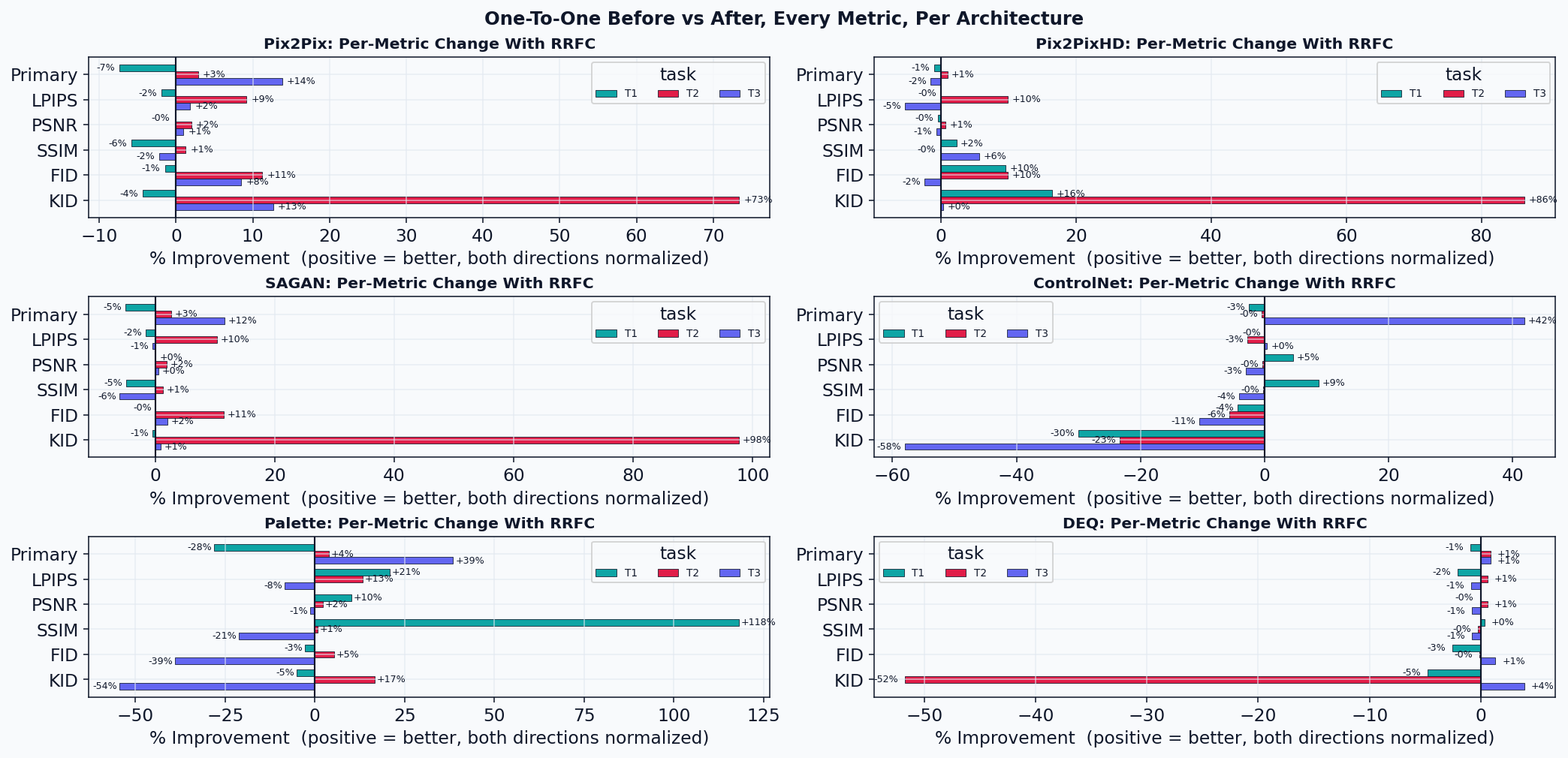}
\caption{One panel per architecture, showing the percentage change with RRFC on every metric (primary, LPIPS, PSNR, SSIM, FID, KID), the three tasks as colored bars, oriented so positive is better. These are relative changes in the seed-averaged metric, so they inflate where the baseline is small (the large KID and identity percentages sit on near-zero baselines and should be read as direction, not absolute effect). On the aligned tasks the companion metrics generally move with the primary, so the benefit is broad; the semantic task again shows SSIM rising while mIoU falls. Being means, a cell's sign can differ from the median forest plot (Figure~\ref{fig:forest}).}
\label{fig:metriccards}
\end{figure*}

\begin{figure*}[tbp]
\centering
\includegraphics[width=\textwidth]{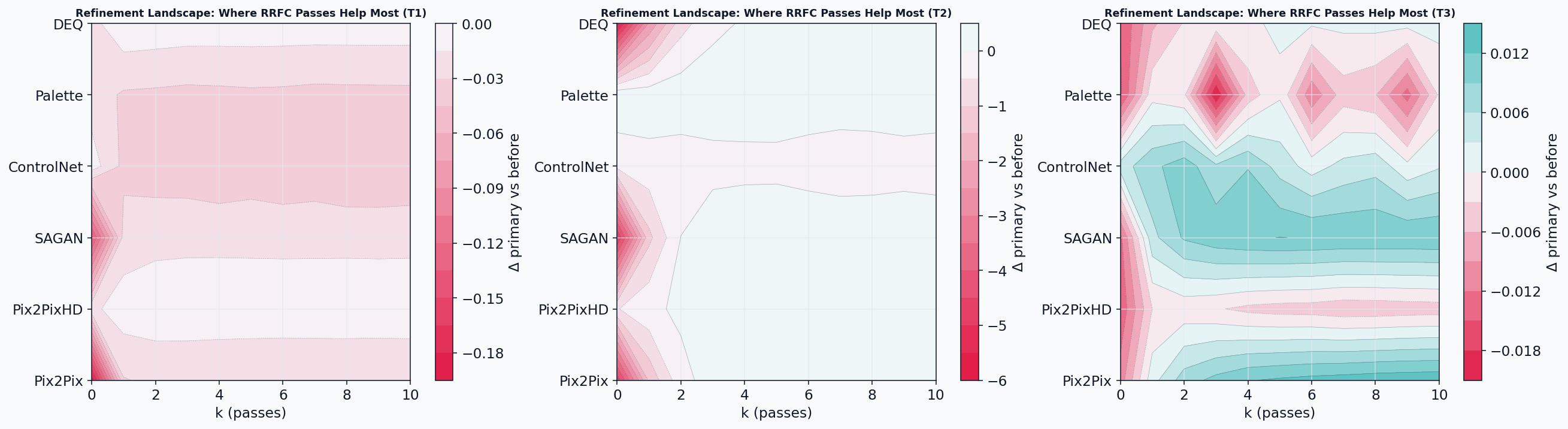}
\caption{Change in each task's primary metric relative to the no-refinement baseline versus refinement depth, averaged over seeds. Each panel is a task, rows are models, the horizontal axis is passes $k$ from 0 to 10, and color is the signed change (teal better, red worse) on a per-panel scale (not comparable across panels). The semantic task (left) is red at every depth; inpainting (middle) is neutral-to-positive across depth; identity (right) turns teal at moderate depth, with isolated red pockets for Palette. Improvements hold roughly flat past the trained unroll depth of three, as the monotonic objective intends.}
\label{fig:refine}
\end{figure*}

\begin{figure*}[tbp]
\centering
\includegraphics[width=\textwidth]{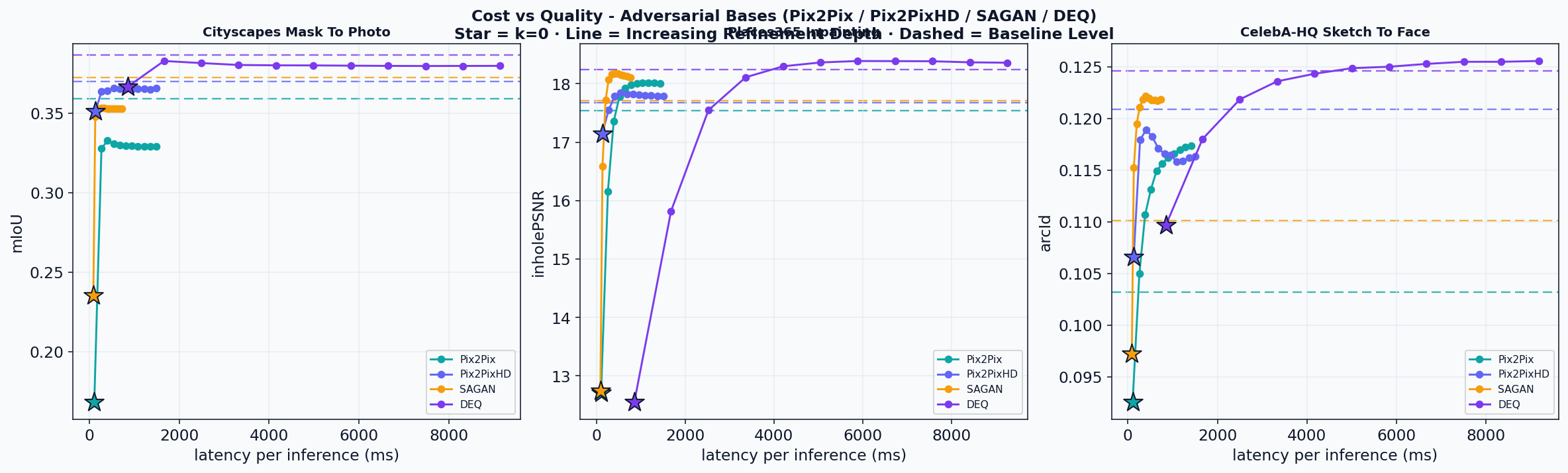}\\[1.5ex]
\includegraphics[width=\textwidth]{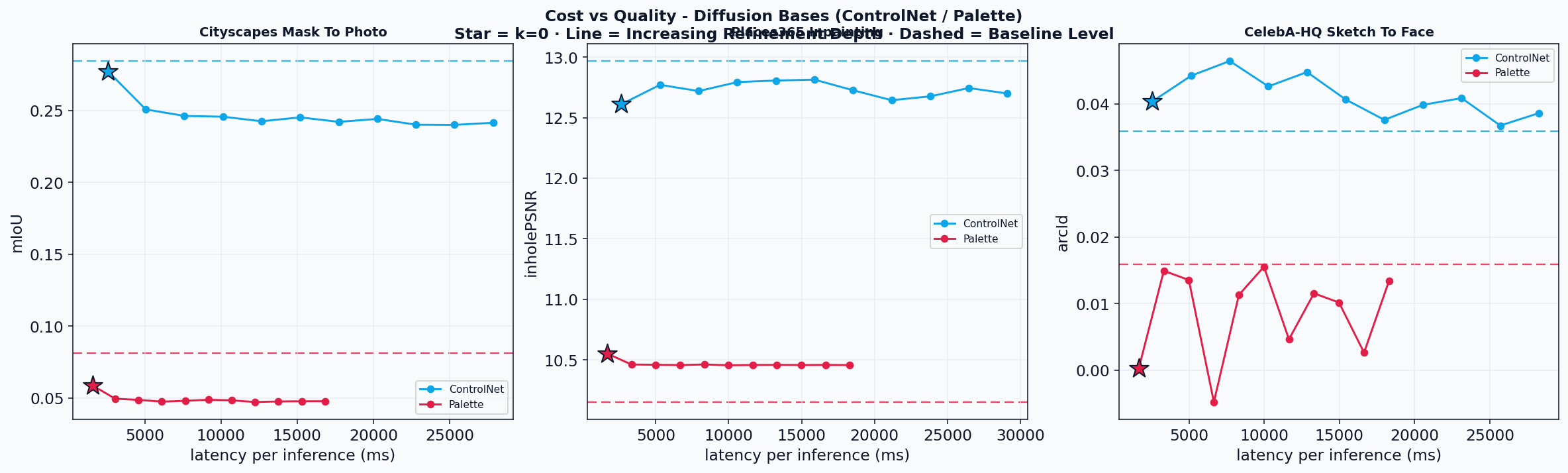}
\caption{Quality against inference latency, stacked with the single-pass GAN and equilibrium bases on top (Pix2Pix, Pix2PixHD, SAGAN, DEQ) and the two diffusion bases below (ControlNet, Palette); each row covers the three tasks. A star marks zero passes, the connected line traces increasing refinement depth, and the dashed line is each model's baseline. The horizontal scales differ sharply between the two rows: the top-row bases reach their useful depth in well under a second per image, whereas the diffusion bases run to tens of seconds, since each of their passes is a full thirty-step DDIM sampler run.}
\label{fig:cost}
\end{figure*}

\begin{figure*}[tbp]
\centering
\includegraphics[width=\textwidth]{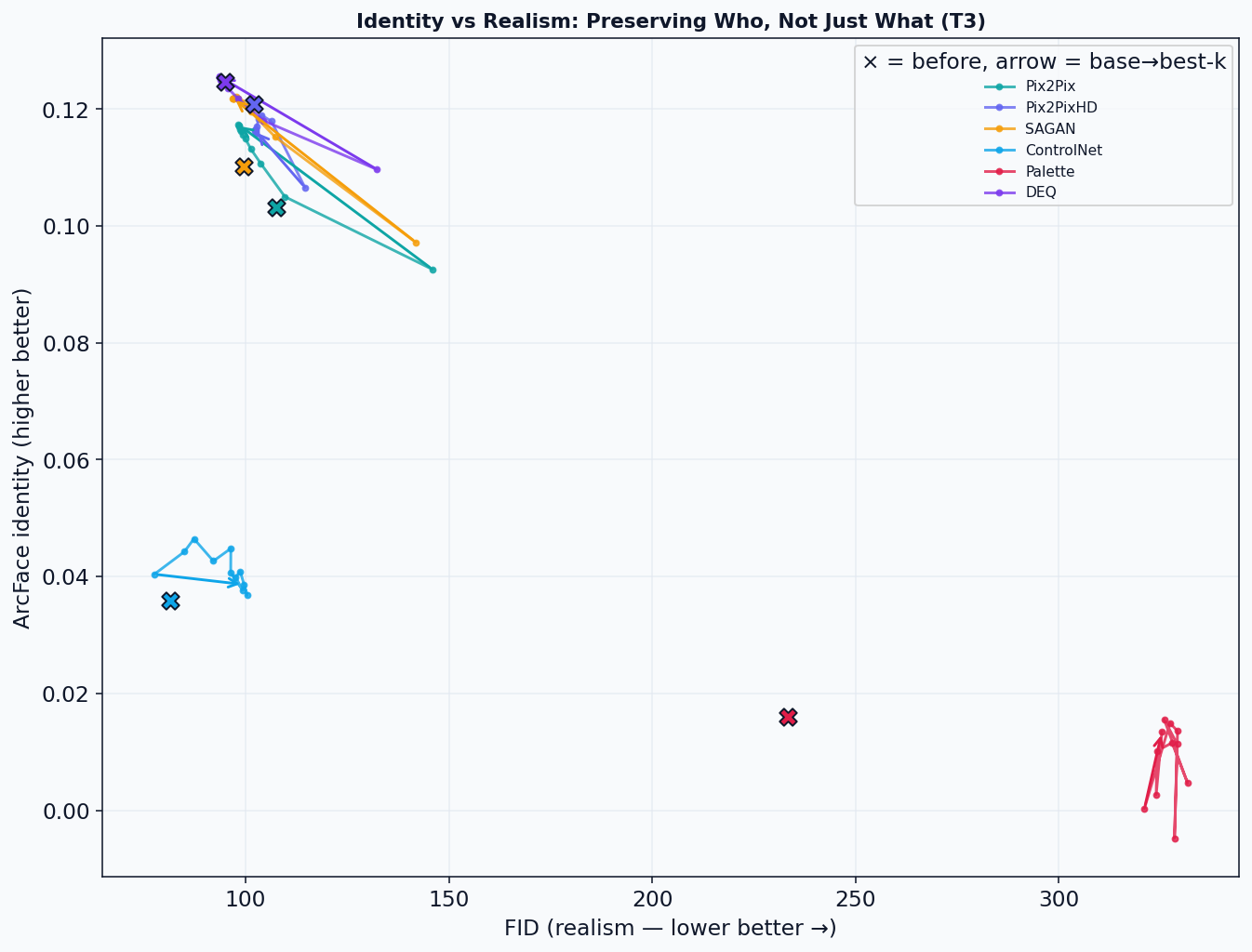}
\caption{Identity against realism on sketch-to-face (T3). The horizontal axis is FID (realism, lower better) and the vertical axis ArcFace identity (higher better); a cross marks each model at zero passes and the arrow runs to its validation-selected best depth. A tight cluster (Pix2Pix, Pix2PixHD, SAGAN, DEQ) sits at low FID and high identity, DEQ and Pix2PixHD strongest near 0.124; ControlNet is isolated at the best realism (FID near 80) but low identity, about 0.04; and Palette collapses to near-zero identity at FID 230 to 330. The two axes are only loosely coupled: a model can be realistic yet fail to preserve identity. Refinement nudges identity up for most models, but the largest relative gains (ControlNet, Palette) sit on near-zero baselines and do not reach the cluster.}
\label{fig:identityrealism}
\end{figure*}

\begin{figure*}[tbp]
\centering
\includegraphics[width=\textwidth]{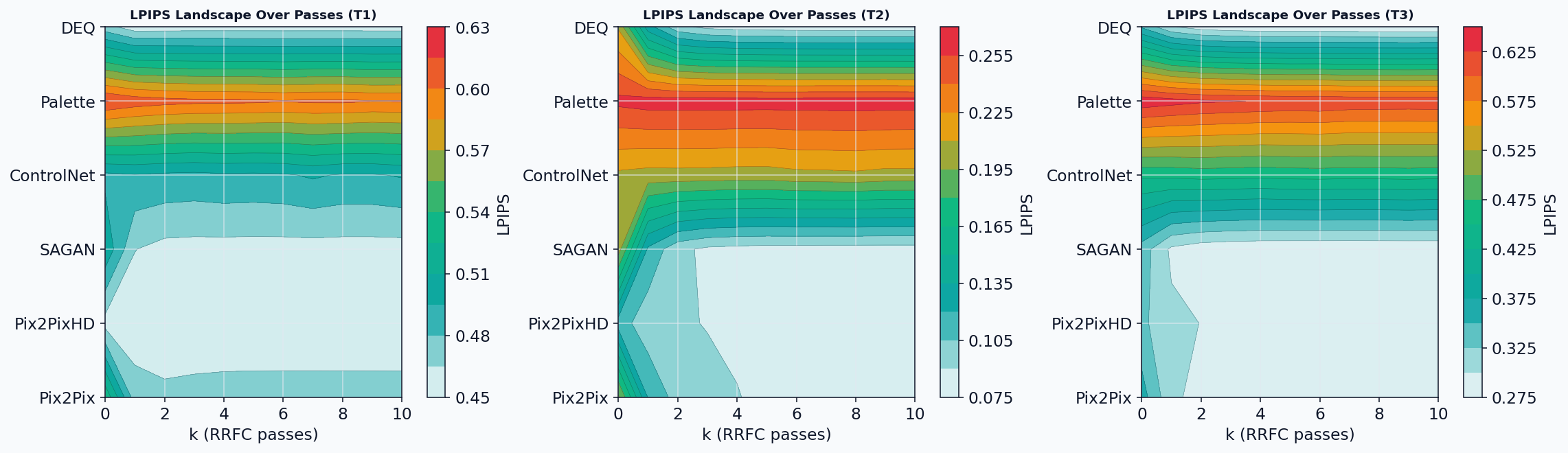}\\[1.5ex]
\includegraphics[width=\textwidth]{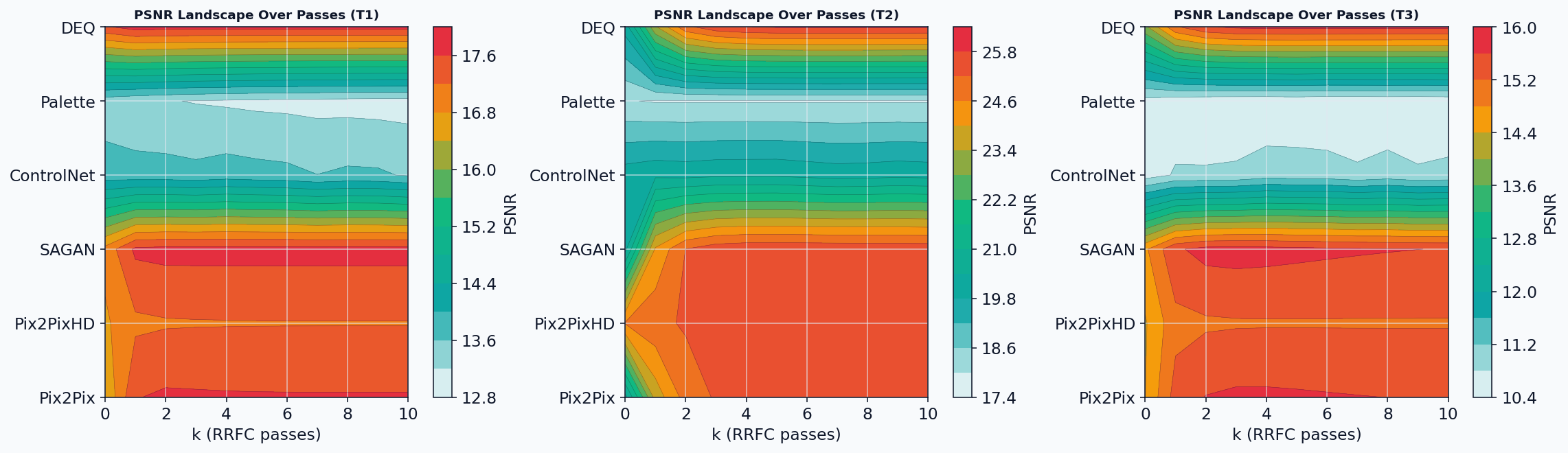}\\[1.5ex]
\includegraphics[width=\textwidth]{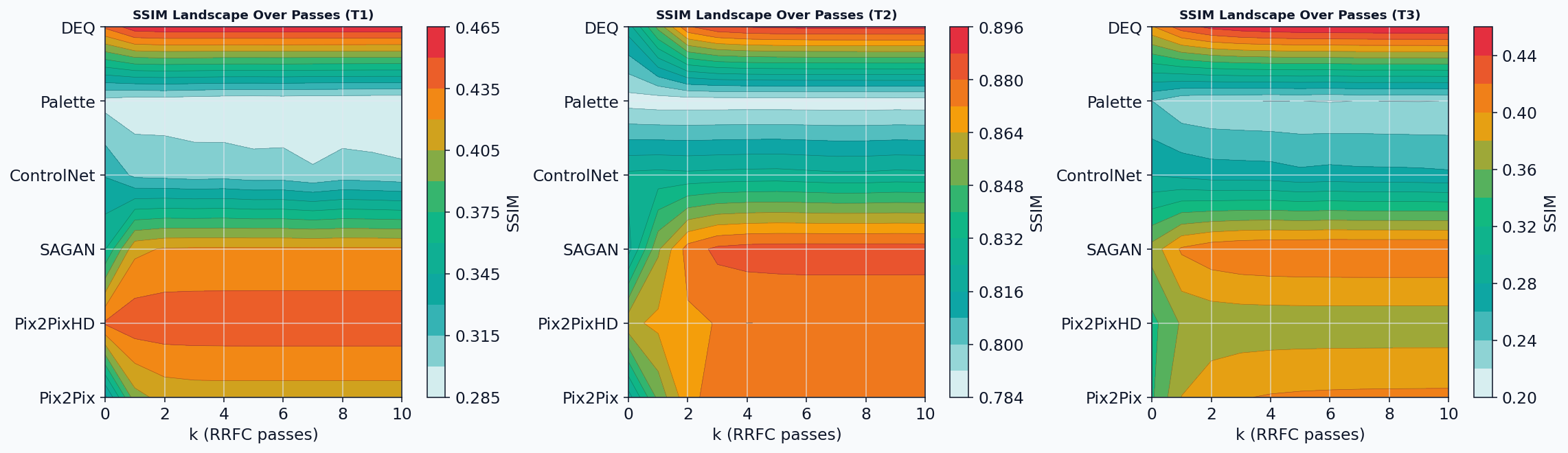}
\caption{Companion-metric landscapes over refinement depth, stacked as LPIPS (top), PSNR (middle), and SSIM (bottom); within each block the three panels are the three tasks, the horizontal axis is passes $k$ from 0 to 10, rows are the six architectures, and color is the metric on a per-panel scale (not comparable across panels or metrics). The variation is mostly by architecture rather than depth: on LPIPS the two Pix2Pix bases are best and Palette worst, on PSNR the single-pass bases hold the highest-fidelity band and Palette is lowest, and SSIM shows the same ordering. Where depth matters, the change is concentrated in the first pass or two and then flattens.}
\label{fig:contours}
\end{figure*}

\begin{figure*}[tbp]
\centering
\includegraphics[width=\textwidth]{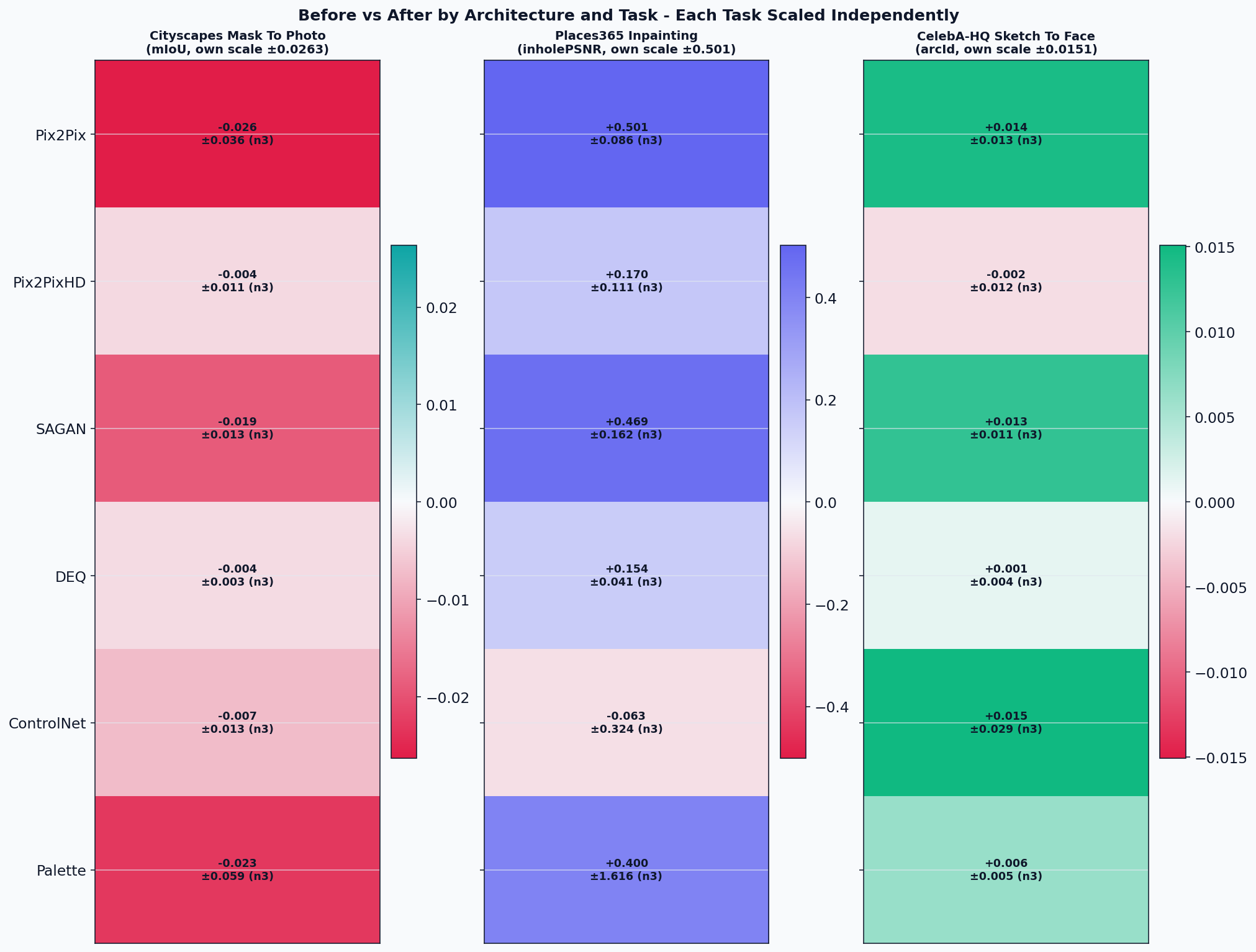}
\caption{Raw before-to-after effect sizes on the primary metric, one column per task each on its own scale, printed as the seed-mean change with its standard deviation over three seeds. The independent scales reflect that the primary metrics are not commensurable (mIoU $\pm$0.026, in-hole PSNR $\pm$0.5 dB, identity $\pm$0.015), so magnitudes should not be compared across columns. The standard deviations matter: several large-looking cells are noise-dominated (Palette on inpainting, $+0.400 \pm 1.616$; ControlNet on inpainting, $-0.063 \pm 0.324$, and on identity, $+0.015 \pm 0.029$), while tight cells such as DEQ are reliable, which is why some non-zero cells fail the Holm test. The signs are negative throughout the semantic column, positive on inpainting apart from ControlNet, and positive on identity apart from Pix2PixHD.}
\label{fig:normheatmap}
\end{figure*}

\begin{figure*}[tbp]
\centering
\includegraphics[width=\textwidth]{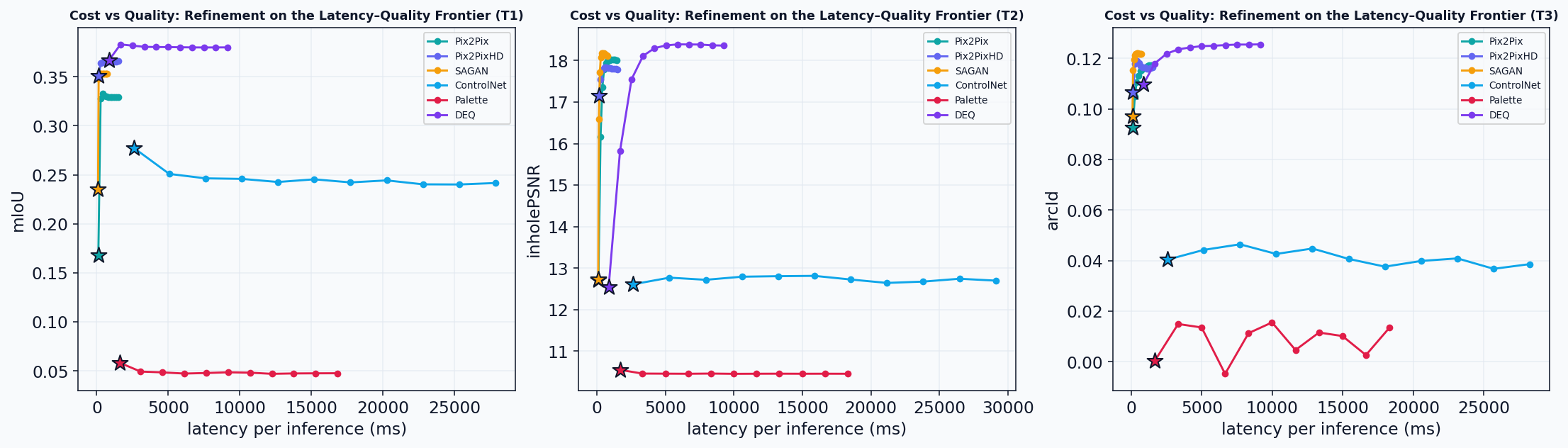}
\caption{The latency-quality frontier with all six bases on a shared axis per task (the architecture-split version is Figure~\ref{fig:cost}); the horizontal axis is wall-clock latency per image in milliseconds, the vertical axis is the task's primary metric, a star marks zero passes, and the connected points trace increasing depth. On one axis the cost gap between architecture classes is immediate: the single-pass bases (Pix2Pix, Pix2PixHD, SAGAN) sit against the left edge at sub-second latency, the equilibrium base DEQ occupies the low thousands of milliseconds, and the two diffusion bases (ControlNet, Palette) stretch to tens of thousands of milliseconds, since each of their passes is a full sampler run.}
\label{fig:costpareto}
\end{figure*}

\end{document}